\documentclass[11pt, letterpaper]{article}
\usepackage[a4paper,top=2.54cm,bottom=2.54cm,left=2.54cm,right=2.54cm,marginparwidth=1.75cm]{geometry}
\usepackage{times}  % DO NOT CHANGE THIS
\usepackage{helvet} % DO NOT CHANGE THIS
\usepackage{courier}  % DO NOT CHANGE THIS
\usepackage[hyphens]{url}  % DO NOT CHANGE THIS
\usepackage{graphicx} % DO NOT CHANGE THIS
\usepackage[authoryear]{natbib}  % DO NOT CHANGE THIS AND DO NOT ADD ANY OPTIONS TO IT
\usepackage{caption} % DO NOT CHANGE THIS AND DO NOT ADD ANY
\usepackage{setspace}
\usepackage{authblk} %

\title{Many Ways to Be Lonely: Fine-Grained Characterization of Loneliness and Its Potential Changes in COVID-19}

\author[1]{Yueyi Jiang\footnote{Corresponding author: yujiang@ucsd.edu}}
\author[2]{Yunfan Jiang}
\author[3]{Liu Leqi}
\author[1]{Piotr Winkielman}
\affil[1]{University of California, San Diego}
\affil[2]{Stanford University}
\affil[3]{Carnegie Mellon University}

\usepackage[table,xcdraw]{xcolor}
\usepackage{amsfonts}
\usepackage{amsmath}
\usepackage{tikz}
\usepackage{subcaption}
\usepackage{dsfont}
\usepackage{booktabs}
\usepackage[export]{adjustbox}
\usepackage[skins]{tcolorbox}

\begin{document}

\maketitle

\begin{abstract}
Loneliness has been associated with negative outcomes for physical and mental health. Understanding how people express and cope with various forms of loneliness is critical for early screening and targeted interventions to reduce loneliness, particularly among vulnerable groups such as young adults. To examine how different forms of loneliness and coping strategies manifest in loneliness self-disclosure, we built a dataset\footnote{The dataset can be found in \url{https://huggingface.co/datasets/FIG-Loneliness/FIG-Loneliness}}, FIG-Loneliness (\textbf{FI}ne-\textbf{G}rained \textbf{Loneliness}) by using Reddit posts in two young adult-focused forums and two loneliness related forums consisting of a diverse age group. We provided annotations by trained human 
annotators for binary and fine-grained loneliness classifications of the posts. Trained on FIG-Loneliness, two BERT-based models were used to understand loneliness forms and authors' coping strategies in these forums. Our binary loneliness classification achieved an accuracy above $97\%$, 
and fine-grained loneliness category classification reached an average accuracy of $77\%$ across all labeled categories. With FIG-Loneliness and model predictions, we found that loneliness expressions in the young adult related forums were distinct from other forums. Those in young adult-focused forums were more likely to express concerns pertaining to peer relationship, and were potentially more sensitive to geographical isolation impacted by the COVID-19 pandemic lockdown. Also, we showed that different forms of loneliness have differential use in coping strategies.
% showed the differences of loneliness discussions across Reddit forums, the relation-ships among loneliness categories, and the effect of pandemic on loneliness.
\end{abstract}

\section{Introduction}
\label{sec:intro}

% This is an example of using color edits.
% \leqicomment{This is a comment that I want to make.}
% \leqiedit{This is a sentence that I want to add.}
% \leqidelete{This is something that I have deleted.}

% \todo{R2: `` Are there any studies which are related to covid-19 and mental health? ''}
% \todo{Mention how young adults are different.} \sherry{done}

% Why loneliness is a pressing issue and why should we understand loneliness?
Feeling socially isolated or lonely is associated with social cognitive impairments, negative health consequences and even mortality \citep{holt2015loneliness, hawkley2010loneliness}. As a result of the COVID-19 pandemic, the “stay-at-home” or “shelter-in-place” orders mandated closure of schools and nonessential businesses, which has resulted in unprecedented change in the amount and type of social interactions, and concerns over increased risk for loneliness, especially among vulnerable groups such as young adults. A recent study conducted during social distancing policies found that loneliness affects young adults more than other age groups \citep{bu2020lonely}. Another study conducted with national samples in the US found (surprisingly) no significant changes in loneliness during the pandemic lockdowns, though confirmed that young adults are especially vulnerable to loneliness \citep{luchetti2020trajectory}. Although sizeable research has examined the prevalence and negative consequences of loneliness, it is little understood how loneliness manifests \citep{mushtaq2014relationship, cacioppo2014evolutionary}.

% how can we understand different manifestations of loneliness and how they fluctuate over time.?
Literature suggests that loneliness is a multidimensional experience, encompassing different manifestations and forms \citep{de1982types, cacioppo2008loneliness}. For example, feelings stemming from the absence of intimate or romantic relationships need to be differentiated from feelings stemming from social isolation in family or friendship contexts. This literature also suggests differences due to interpretation of the situation as changeable or chronic. It is crucial to differentiate among forms of loneliness (e.g., romantic versus social loneliness, situational versus chronic loneliness) as research suggests that different sources of loneliness are associated with different forms of psychopathology \citep{Lonelinesspsychopathology}. Of different forms of loneliness, chronic loneliness should be given special attention because of its potential harmful consequences in mental health including increased risks for depression and suicidality \citep{perlman1984loneliness}.

Notably, the persons' coping strategies for loneliness are important to examine. Active coping strategies, sometimes called problem-focused strategies, encourage the person to pursue solutions to alleviate loneliness, whereas passive coping strategies or emotion-focused strategies focus on eliminating the negative feelings associated with the source of stress. Previous studies demonstrate that active coping styles were associated with lower levels of loneliness compared to passive coping styles \citep{deckx2018systematic}. 
%Based on a review of previous work, 
% It is possible that those who express situational loneliness will incorporate more active coping strategies using active interaction seeking as a proxy than chronic loneliness. 
%\leqicomment{Did we later on confirm this hypothesis?}
Note also that loneliness manifestations can fluctuate over time. The impact of the COVID-19 pandemic (a time when the forms and amount of social interactions are changed) on different loneliness manifestations remains to be answered. 

% Recent technologies uses social media data to facilitate the understanding of loneliness.
Previous research has suggested that understanding online disclosure can facilitate early detection of psychological challenges (see review in \citet{chancellor2020methods}). Reflecting the wide adoption of social network platforms (e.g., Twitter, Instagram and Reddit) that enable individuals to share their emotional and social experiences publicly, 
online disclosure not only becomes an important therapeutic ingredient in improving psychological well-being \citep{ellis2012emotional, turner1983social}, but also provides researchers opportunities to understand naturalistic patterns of mental states \cite{olteanu2017distilling}. Given that social media use has been rapidly growing and young adults constitute the majority of users \citep{socialmediademographics}, this raises an interesting question: how do people, especially young adults, use social media platforms to express or cope with their loneliness experiences? 

Some answers were found in recent studies that examined the impacts of COVID-19 on online mental health discourse. \citet{low2020natural} showed that the number of posts on Reddit mental health forums mentioning suicidality and loneliness-related topics was more than doubled. Another study examined the themes using tweets containing co-occurred words ``lonely'' and ``COVID-19'' \citep{koh2020loneliness}. They found that discussions on loneliness during the pandemic focused more on mental health effects of loneliness rather than community impact of loneliness from May to July 2020. Note though that these studies treated loneliness as a broad, unitary, and explicit theme.  They used loneliness related keywords or forum memberships for posts' ground truth labels for loneliness and did not differentiate types, duration, and contexts of loneliness.  Consequently, their selection method does not allow for other loneliness expressions with no explicit mentions of such keywords, and can result in underreporting due to the social stigma of admitting to feeling lonely \cite{rokach1997loneliness}.

% Additionally, although previous research looked at changes in language over a one-week period in loneliness disclosures \citep{mahoney2019feeling} , the pattern of loneliness expressions over a large time scale has not been fully explored. Thus, an important question to be answered is how different types of loneliness expressed in online discourse change over time, especially after social distancing due to the COVID-19 pandemic?

% \todo{Modify. Also make it clear that we want to distinguish general loneliness v.s. young adult loneliness.
% ``Narrative and scope: R1 and R3 recommended to scope the narrative of the paper to make a narrower but clearer contribution. R1 also sought further methodological validation with respect to the robustness of the claims and results of the paper.''} \sherry{done}

% In this study, we aim to identify online loneliness discourse over time on Reddit, and provide fine-grained characterization of different types of loneliness as suggested in psychological theories. We have three hypotheses:
In this study, we aim to provide \emph{fine-grained} characterization of loneliness discourse including the forms of loneliness and coping strategies for loneliness as suggested in psychological theories and literature. To complement the selection method described above, we captured a wider range of loneliness expressions by using both human annotations and model predictions. We set out to examine three main research questions:
\begin{itemize}
    \item Loneliness expressions: How do people express loneliness experiences in online communities? (Section~\ref{sec:rq1}) %``How do young adults focused communities and communities with diverse age groups express loneliness?''
    \item Coping strategies: Are different forms of loneliness associated with different coping styles? (Section~\ref{sec:rq2})%(RQ 2) 
    %``How do different forms of loneliness associated with the types of coping strategies?''
    \item Impacts of the COVID-19 pandemic: How does the pandemic affect loneliness discourse? (Section~\ref{sec:rq3})%(RQ 3) 
    %``How do COVID-19 impact different type of loneliness on Reddit?''
\end{itemize}

In the analyses of loneliness expressions and the impacts of the pandemic, we will draw inferences about loneliness characterization from young adult-focused communities and communities of diverse age groups separately. We used Reddit to examine these questions. Reddit is a social media platform where people communicate on topic-specific forums, called subreddit. This platform can capture language used by people disclosing their pandemic experiences on public online communities in real-time. Because user accounts are anonymous, Reddit has also been widely used to investigate self-disclosure in mental illnesses \citep{de2014mental}. In addition, a Reddit post has a limit of $40,000$ characters, compared to a Tweet which has a $280$-character limit and an Instagram status post which has a limit of $2,200$ characters. Thus, drawing data from Reddit provides opportunities for more comprehensive content analyses. 

Taking both qualitative and quantitative approaches, we characterize discussions about loneliness experience on Reddit forums and provide strong models for fine-grained loneliness classification. Specifically, we manually annotated thousands of Reddit posts that discussed loneliness experience across years (2018, 2019, and 2020). Using the labels from both human annotations and model predictions, we investigated how people used Reddit forums for different forms of loneliness expressions across different age groups, the relationship between loneliness forms and coping strategies, and explored how loneliness discussions changed since the COVID-19 pandemic. 
Our main contributions are three-folds:
\begin{enumerate}
    \item We built a large human annotated dataset, FIG-Loneliness (\textbf{FI}ne-\textbf{G}rained \textbf{Loneliness}), that is suitable for content analysis on fine-grained loneliness (Section~\ref{sec:dataset});
    \item Using hierarchical distributional learning, we built strong classifiers to identify different forms of loneliness and coping strategies within online expressions (Section~\ref{sec:methodology});
    \item Using FIG-Loneliness and the obtained classifiers, 
    we examined the proposed research questions with a focus on loneliness discourse from young adults, which is a group at greater risk for loneliness
    (Section~\ref{sec:research-questions}). 
    %\leqicomment{Maybe we could just say to examine the three aforementioned questions on loneliness expressions, coping strategies and the effects of the COVID-19 pandemic? Or maybe we should somehow say something in addition to ``young adults.''}
\end{enumerate}
Our study provides important insights into how loneliness experiences manifest in online discourses, 
which is not only helpful for improving loneliness screening tools but also for designing effective and targeted loneliness interventions for vulnerable groups, such as young adults.
%In summary, the novel contributions of this project are as follows.
%1. We built the largest dataset containing loneliness fine grained labels 
%2. application of multi-label classification
%3. we characterized the types of loneliness discussed on online Reddit forums across time

\section{Related Work}
\label{sec:related-work}
The current paper is motivated by the need to examine online discourses around mental health concerns through social media data, which we describe in Section~\ref{2.1}. Our study is most related to two categories of work: classifying loneliness and characterizing loneliness expressions in social media. The former category can be divided into two kinds of classification problems, which are the predictions of an expression being lonely or non-lonely, and the specific forms of loneliness it expresses. The latter category is related to characterization of loneliness. We describe our loneliness classification scheme and summarize the supporting research and theories in Section ~\ref{2.2} and Section~\ref{2.3}.

\subsection{Self-disclosure in Online Mental Health Communities} 
\label{2.1}

Due to the lack of access to mental health services, individuals with mental health concerns are turning to online mental health communities such as Reddit to share their emotional challenges and seek social support. The goal of such communities is to provide a "safe haven" for mental health disclosure and peer-to-peer support for stigmatized concerns \cite{saha2020understanding}. Previous research suggests that self-disclosure through anonymous communications may promote better mental health outcomes. For example, Andalibi et al. studied self-disclosure of sexual abuse on Reddit and found that those who used a more private means of communication like "throwaway" accounts engaged more in seeking support than those who used identifiable accounts \cite{andalibi2016understanding}. Similarly, Yang et al. investigated how the use of private and public affected members' self-disclosure in an online health support group and  found that negative self-disclosure in the private channels was associated more with receiving both informational and emotional support compared to the public channels \cite{yang2019channel}.

As online mental health communities have become a promising venue for studying self-disclosure related to mental health concerns, researchers have used this venue to detect the presence of major depression, suicidality, schizophrenia and other mental health problems \cite{chancellor2020methods}. Compared to the above-mentioned mental health concerns, loneliness, a pervasive and growing public health concern \citep{holt2015loneliness}, has received relatively little attention. This paper fills in an important gap by focusing on understanding and predicting different forms of loneliness experiences in the context of online self-disclosure.

% Although there has been considerable interest in 

% Sharma and De Choudhury built classification models to quantify the forms of online social support in Reddit \cite{sharma2018mental}. Specifically, they focused on emotional support such as empathetic messages or informational support such as advice regarding mental health treatment. Relatedly, Sharma et al emphasized the need for empathy training to provide better support for 

\subsection{Loneliness Classification}
\label{2.2}
% \todo{More related work on classify loneliness. Content analysis, e.g., social media posts and then human annotator to label; simpler methods, e.g., keyword}
Current research on loneliness classification is limited. Loneliness has often been reported as a negative emotion emerged from discussions surrounding mental health on Reddit~\citep{de2014mental, low2020natural}. 
\citet{guntuku2019studying} built a Random Forests Classifier by using extracted linguistic features from tweets to predict mentions of loneliness that include the words ``lonely'' or ``alone''. It should be noted that this does not include expressions without the mentions of loneliness keywords. Instead of using keyword filters as ground truth metrics for loneliness, we provide a human annotated dataset that includes posts without mentions of loneliness related keywords. This avoids prediction bias for keywords when training a classifier and captures a wider range of loneliness expressions.

% It should be noted that the trained dataset for our model includes posts that \emph{do not} contain loneliness related keywords, which avoids prediction bias for keywords and enables a robust classification for a wider range of loneliness expressions.

 One great challenge is that the nature of loneliness is multifaceted. 
 To our best knowledge, %To the best of our knowledge, 
 there has not been research on building predictive models to classify different forms of loneliness. Thus, a question emerges: How to leverage machine learning tools to understand different forms of loneliness in social media discourse? We aim to close this gap by systematically examining the forms of loneliness presented in online disclosure, and building classifiers on our dataset that contains fine-grained loneliness annotations. Our classification scheme enables us to explore the composition of different forms of loneliness in online loneliness expressions across different groups (e.g., by age demographics), and examine the relationship between specific loneliness forms and users' coping strategies.
 We provide more details on our fine-grained loneliness characterization below.
 %We will describe our fine-grained loneliness classification scheme in more detail below.

%on Reddit.
%\leqi{I think it would be great to add an example here where keyword is not a good indicator for lonliness.} 

\subsection{Characterization of Loneliness Expressions}
\label{2.3}

% \leqi{Maybe: Category of Loneliness Expressions?}}
\label{subsec:loneliness-characterization}
The most relevant investigations to our study have focused on characterizing loneliness expressions using tweets. \citet{mahoney2019feeling} captured loneliness discourse on Twitter containing a single term ``lonely'', and categorized the posts into themes. Another study conducted by \citet{kivran2014understanding} qualitatively categorized loneliness expressed on Twitter in three dimensions: (i) the temporal bonding of loneliness; (ii) the context of loneliness; and (iii) interactivity (interaction with others or not) within the expression. In their study, tweets containing specific phrases such as ``I'm so lonely'' were selected for characterization. Both studies could potentially exclude other loneliness expressions outside the curated phrases. 

% Additionally, although the former study looked at changes in language over a one-week period in loneliness disclosures, the pattern of loneliness expressions over a large time scale has not been fully explored.

Expanding upon~\citet{kivran2014understanding}, we focus on four categories 
% aspects \leqicomment{Do we want to change ``categories'' below to ``aspects''?} 
of loneliness disclosure: \emph{duration of loneliness}, \emph{contexts of loneliness}, \emph{interpersonal relationships} involved in loneliness and \emph{interaction styles} {(including coping strategies)} of the disclosure. 
%\leqicomment{Currently, when we introduce interaction in Section 3, we focused on interaction styles instead of interaction intent?}
We are particularly interested in these aspects because previous research noted that the temporality of loneliness and the strategies to cope with loneliness are associated with differential mental health outcomes \citep{perlman1984loneliness,deckx2018systematic}. %(see research summary in Section~\ref{sec:intro}).
Moreover, different types of contexts and relationships meet different needs for social connection, and thus can contribute {as} different sources of loneliness \cite{cacioppo2008loneliness}. 

In our loneliness classification scheme, we annotated a given loneliness expression for different {forms} of loneliness. Specifically, the duration of loneliness refers to whether the loneliness experience is a transient ``state'' or a chronic ``trait'' \cite{perlman1984loneliness}. 
% Note that we cannot retrieve information about the exact timeline of the loneliness experience from the posts, so we will refer to transient and enduring loneliness in this study. 
The contexts of loneliness cover social (relationships with others), physical (the impact of environment change), somatic (physical or bodily states), and romantic (romantic relationship) domains, which are defined similarly to the coding scheme used in \citet{kivran2014understanding}. We further categorized the types of interpersonal relationships mentioned in the post into friendship, family, peers (classmates or colleagues), and romantic, which are not mutually exclusive from the contexts. This is similar to the Differential Loneliness Scale in~\citet{schmidt1983measuring} which measures dissatisfaction with four types of relationships: family, group or community, friendships, and romantic/sexual relationships. 

Finally, we used active and passive interaction seeking as a proxy for problem-focused and emotion-focused coping strategies respectively for loneliness. We classified the posts into five interaction strategies: (i) seeking advice such as checking if social rules are followed (e.g., ``is it okay for me to tag along''); (ii) seeking affirmation and validation (e.g., ``am I doing the right thing?''); (iii) reaching out (e.g., ``want to talk to someone''); (iv) providing social support to combat loneliness\footnote{Some users shared their own loneliness experiences and coping strategies to help others, which could reflect community resilience.} (e.g., ``I joined a book club and feel less lonely now''); and (v) non-directed interaction such as venting and storytelling with no desire for interactions (e.g., ``I'm ugly and will die alone''). We determined (i)-(iii) to represent the problem-focused (active) coping strategies and (v) to represent emotion-focused (passive) coping strategies.

% This is consistent with the evolutionary approach proposed by~\citet{cacioppo2008loneliness} in which they argue that loneliness constitutes an aversive internal state that motivates (i.e., active interaction seeking) individuals to attend to social relationships, which can aid cooperation and self-protection \cite{layden2018loneliness}.

\section{FIG-Loneliness : A Dataset for Fine-grained Loneliness Characterization}
\label{sec:dataset}
Below we describe how we constructed FIG-Loneliness, a dataset for fine-grained loneliness characterization and subsequent model training. First, by using the Reddit's Pushshift API  \citep{baumgartner2020pushshift}, we collected all posts from two loneliness specific subreddits (\texttt{r/loneliness, r/lonely}) and two subreddits for young adults (\texttt{r/youngadults, r/college}) from 2018 to 2020. The idea behind this design is to get data on loneliness expressions not only from a wider user base but also from users who specifically belong to the young adult age group.
Subreddits \texttt{r/lonely} and \texttt{r/loneliness} have over $244,000$ forum members and have a focus on discussions surrounding loneliness-related issues. 
For example, the homepage of \texttt{r/loneliness} sends a message that ``if you are lonely enough to enter /r/loneliness in the address bar like I did, just to see if such a reddit exists, this might be the reddit for you. Say hello!".
% The description of one forum sends a message that "if you are lonely enough to enter /r/loneliness in the address bar like I did, just to see if such a reddit exists, this might be the reddit for you." 
Thus, we assumed that both subreddits are communities for people who experience loneliness to post and connect, and contain users from diverse age groups. We also chose two non-loneliness specific subreddits (\texttt{r/youngadults, r/college}) which are created for discussing a variety of topics with a community of young adults or college-age students.
% \todo{Add \texttt{r/youngadults}.
% R2: ``section 3: Why these three sub-reddits are selected? It is not clear why only r/college is studied and assumed to be between the age 18 to 25. Can people in r/loneliness fall outside college group age range?'' 
% } \sherry{done}
After we removed posts that were deleted by the original posters or repeated entries and only retained English posts,
our dataset before annotation includes $84,639$ posts from \texttt{r/lonely}, $3,382$ posts from \texttt{r/loneliness}, $3,689$ posts from \texttt{r/youngadults} and $101,751$ posts from \texttt{r/college}. To ensure meaningful content for annotation, we only consider posts with at least 25 words in the post title and body combined.% \todo{yunfan: double check the post numbers} 

Next, we sampled data for annotations and model training. The posts used for fine-grained loneliness annotation were randomly sampled from all four subreddits. For our lonely samples, we consider all posts from the loneliness-related subreddits and only the posts that contain loneliness-related keywords (i.e., alone, lonely, lonesome, loner, loneli, loneness, isolated and left out) in \texttt{r/youngadults} and \texttt{r/college}. With this selection method, we included posts without the explicitly mentioned loneliness keywords. For our non-lonely samples, we randomly selected posts from \texttt{r/youngadults} and \texttt{r/college} that do not contain such keywords. These posts potentially do not contain loneliness expressions. The selected lonely and non-lonely samples constitute the basis of the dataset for annotation. To ensure that the sampled data is representative of the total data, we kept both ratios among the four subreddits and ratios between pre-pandemic (2018, 2019) and post-pandemic (2020) for the samples consistent with the total data.

% We then constructed a dataset for lonely and non-lonely posts separately. For the lonely dataset, we first filtered the posts using loneliness related keywords (i.e., alone, lonely, lonesome, loner, loneli, loneness, isolated and left out). We randomly sampled 3000 posts from both pre-pandemic (2018, 2019) and post-pandemic (2020) periods from the three subreddits. We kept the fractions of the samples from the subreddits the same as the proportions of these subreddits. The non-lonely dataset contains posts without any loneliness filter words from the college subreddit. 

\begin{figure}[ht!]
  \centering
  \includegraphics[width=\linewidth, center]{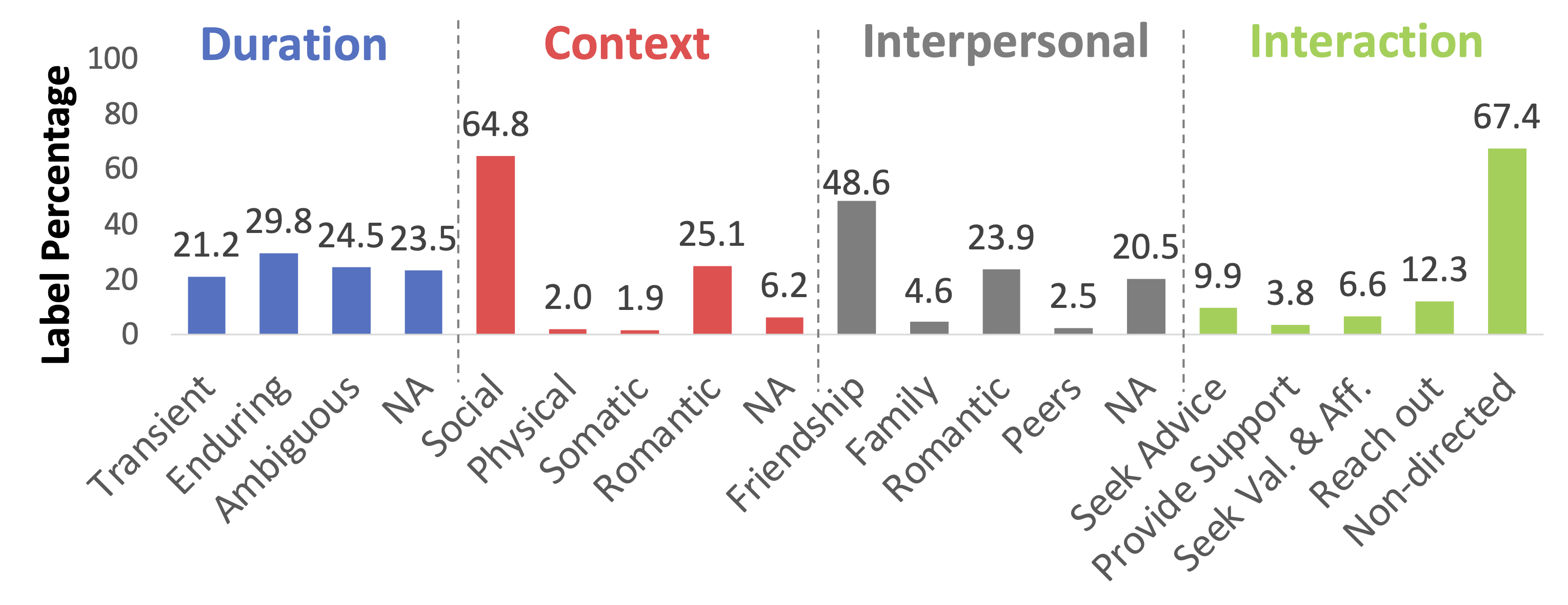}
  \caption{Distributions of labels for annotated lonely posts. %Label frequency of each category in the annotated dataset is shown.
  }
  \label{fig:label_freq}
\end{figure}

To provide high-quality fine-grained ground truth labels for the selected posts, we had both trained undergraduate research assistants and Amazon’s Mechanical Turk workers (MTurkers) with a Master certification provide annotation labels. Six research assistants labeled the sampled potential lonely posts. Each post was labeled by three of them.
%we created a labeled dataset by annotating each post with six annotators. 
A posts was first labeled on whether it contains an expression on self-disclosure of loneliness. If the majority of the annotators labeled a post as not containing such expression, the post would be discarded, otherwise it is further labeled according to a codebook that contains the following categories: 
(1) \emph{duration}: the duration of the loneliness experience (transient, enduring, and ambiguous\footnote{``Ambiguous loneliness'': annotators cannot determine if a post belongs to enduring or transient loneliness.}), (2) \emph{context}: the contexts of the experience (social, physical, somatic, and romantic), (3) \emph{interpresonal}: the interpersonal relationships involved in the experience (romantic, family, friendship, and peers), and (4) \emph{interaction}: user interaction styles (seeking advice, providing support, seeking validation/affirmation, reaching out and non-directed interaction). The codebook is intended for dissecting different {forms} of loneliness %(categories (1)-(3)) 
and users' coping strategies  %(category (4)) 
in the loneliness discourse. 
We also included a ‘not applicable’ (NA) label to accommodate situations that are not suitable for {classification}. For each category, the annotators gave one value which they think would best capture the source of loneliness in the post or the author's interaction intent. The hierarchical labeling structure of a post is illustrated in Figure \ref{fig:trees}. 
We explained the details of each label and the choice rationale in Section~\ref{subsec:loneliness-characterization}.
% For {more explanations} on the constitution of each category \leqiedit{and why these categories and labels are chosen}, 
% we refer the readers to Section~\ref{subsec:loneliness-characterization}.
The distributions of labels for the annotated lonely posts are shown in  Figure~\ref{fig:label_freq}. 
For the potential non-lonely samples, MTurkers were instructed to classify whether the posters express loneliness. Each post was annotated by three MTurkers, and only posts labeled as non-lonely by the majority would remain in the final annotated dataset. 

\begin{figure}[ht!]
    \centering
    % \begin{subfigure}{0.45\columnwidth}
    %   \centering
    %   \includegraphics[scale=0.33]{figures/inter_rater_agreement_lonely.pdf}
    %   \caption{Loneliness}
    % \end{subfigure}
   % \hfill
    \begin{subfigure}{0.45\columnwidth}
      \centering
      \includegraphics[scale=0.33]{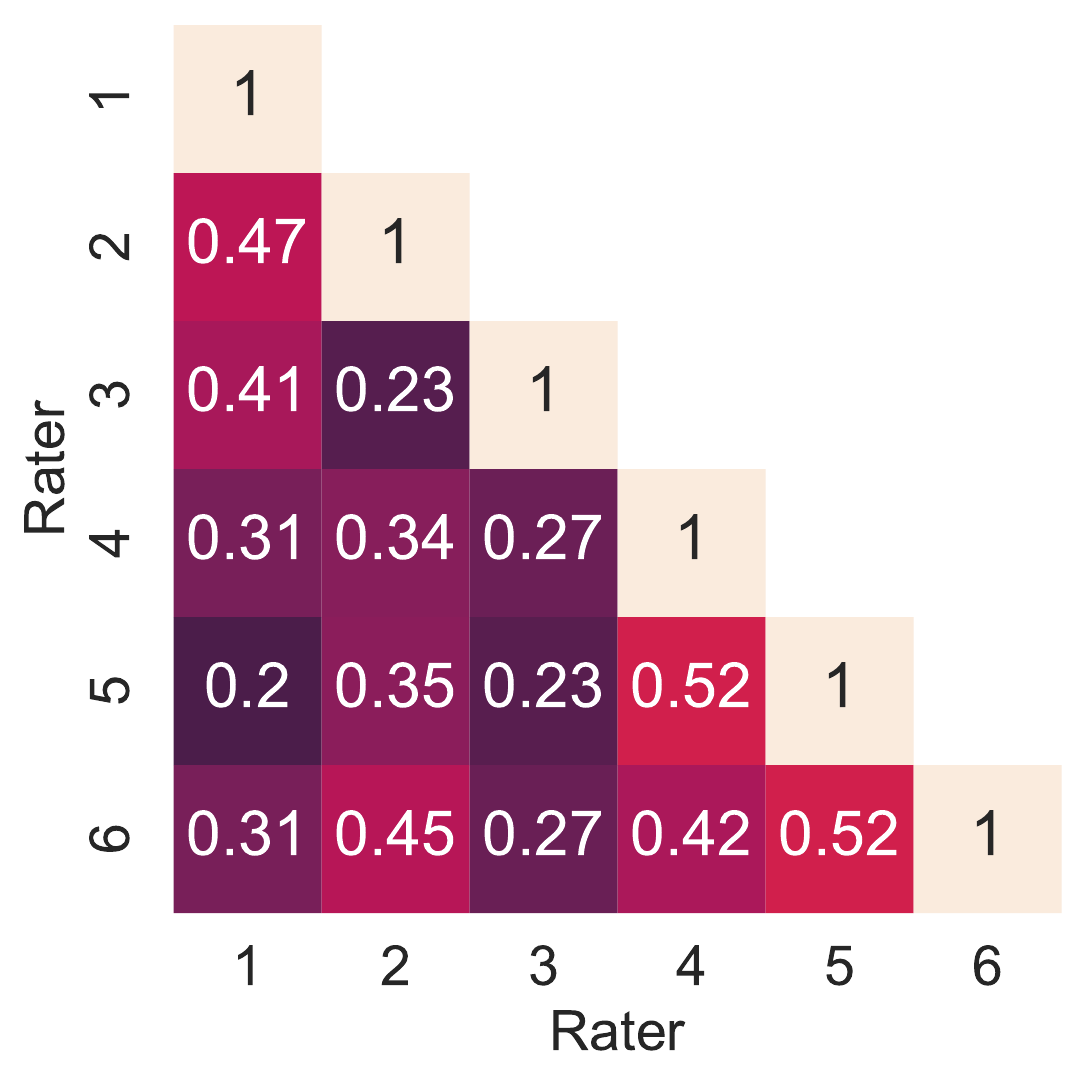}
      \caption{Duration}
      \label{fig:inter-rater-duration}
    \end{subfigure}
    \hfill
    \begin{subfigure}{0.45\columnwidth}
      \centering
      \includegraphics[scale=0.33]{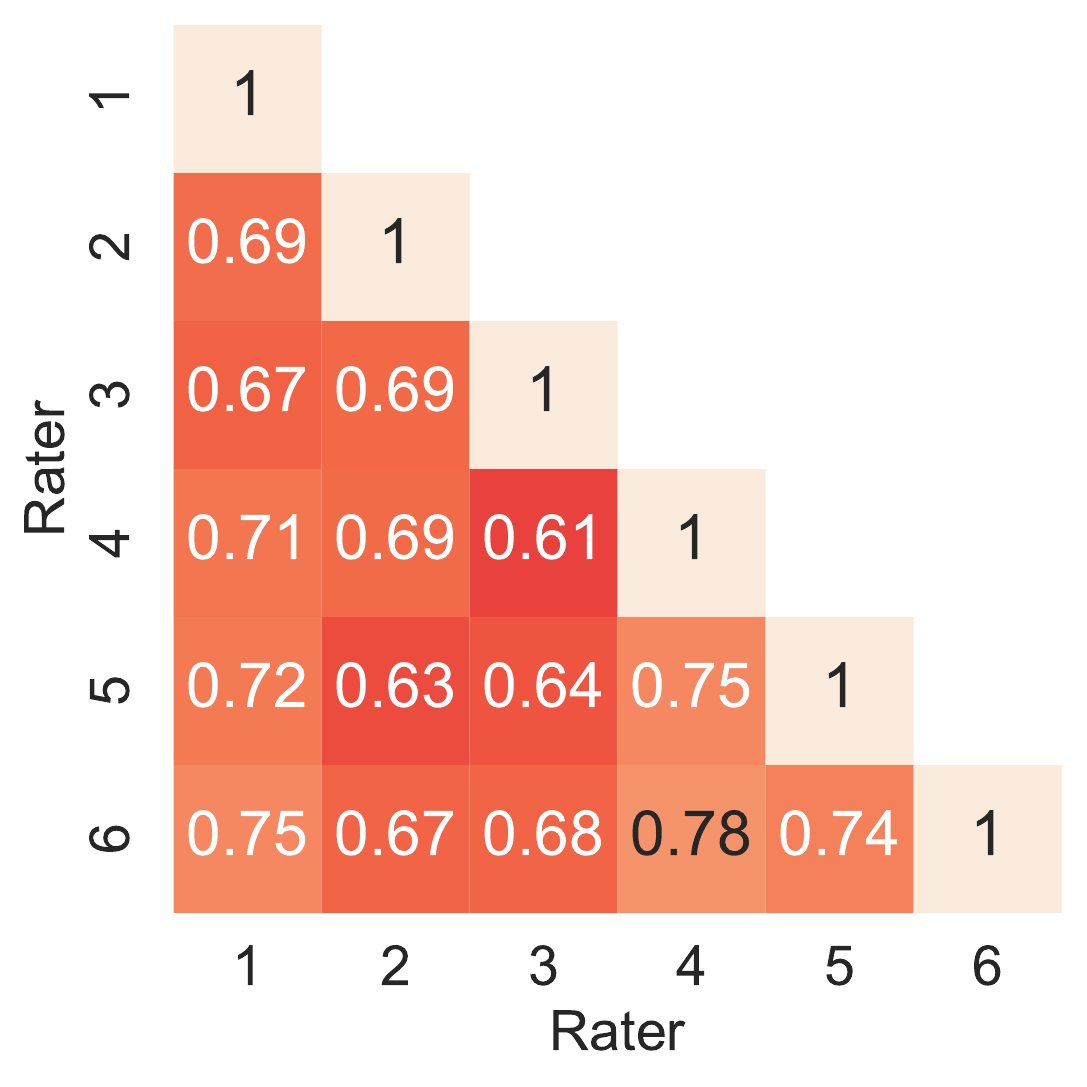}
      \caption{Context}
    \end{subfigure}
    \hfill
    \begin{subfigure}{0.45\columnwidth}
      \centering
      \includegraphics[scale=0.33]{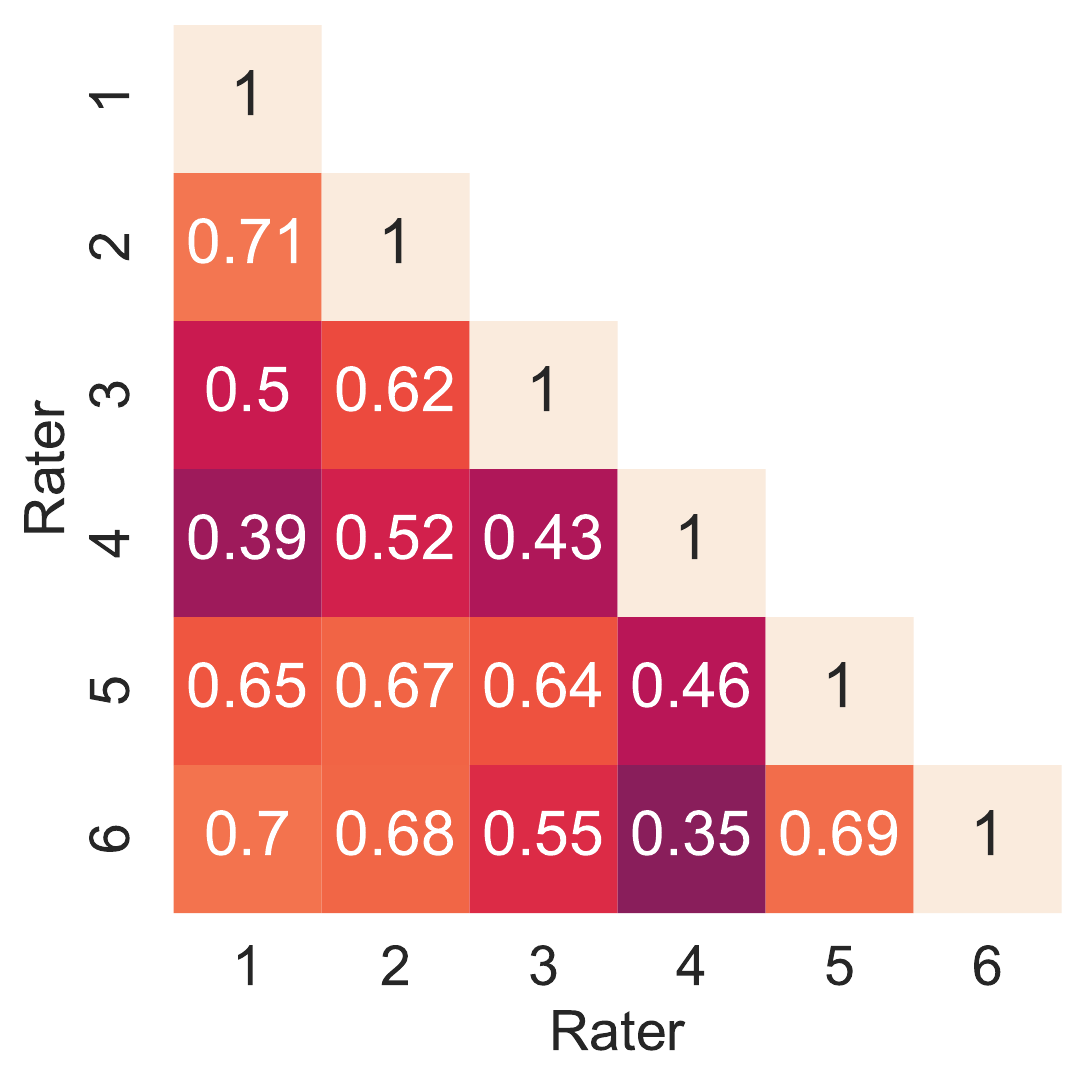}
      \caption{Interpersonal}
    \end{subfigure}
        \hfill
    \begin{subfigure}{0.45\columnwidth}
      \centering
      \includegraphics[scale=0.33]{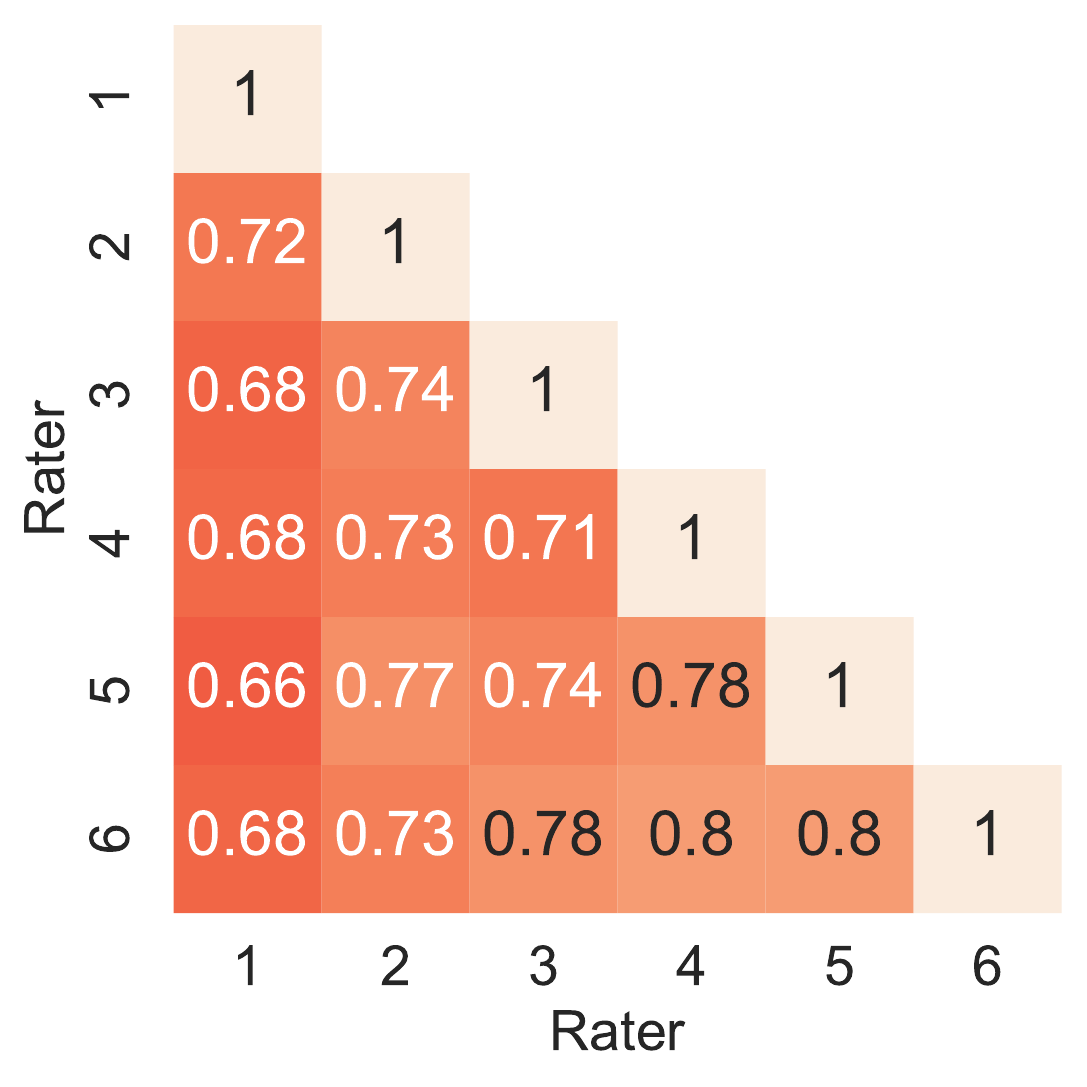}
      \caption{Interaction}
    \end{subfigure}
    \caption{Inter-rater agreements per category. Values range from $0$ to $1$, with higher values indicating greater similarity two raters' labels share.}
    
    % \yunfan{Can we say all loneliness annotations agree so we dont need the first subfigure?} 
    % \yunfan{DONE}
    % \todo{Change the similarities to per category similarity. Add a sentence to say the higher, the better. Range between 0 and 1.}
    
    \label{fig:inter_rater}
\end{figure}

All the labeled posts and annotations were included in FIG-Loneliness, which consists of roughly 3000 lonely and 3000 non-lonely posts.
Finally, FIG-Loneliness also includes other information including the posters' demographics (e.g, gender, age and occupation) and mental health states if mentioned in the post. Consistent with our assumption, as shown in Table~\ref{demographics}, the age groups from \texttt{r/lonely} and \texttt{r/loneliness} are more diverse than \texttt{r/youngadults} and \texttt{r/college} in the annotations for age.
\begin{table*}
\centering
  \linespread{0.7}\selectfont\centering
\begin{tabular}{@{}ccccc@{}}
\toprule
                           &\% of lonely posts with age labels  & before 18 & 18 - 25 & 25+    \\ \midrule
\texttt{r/youngadults} \& \texttt{r/college} & 28.6\% & 16.7\%    & 83.3\%  & 0      \\
\texttt{r/lonely} \& \texttt{r/loneliness}  & 24.5\% & 24.9\%    & 47.8\%  & 27.3\% \\ \bottomrule
\end{tabular}
\caption{Age demographics of annotated lonely posts. This is an estimate of our data representation as not all annotated posts contain age information.
The last three columns show the percentage of different age groups among lonely posts with age labels.} %\sherry{we can also show "n" for the two groups' age annotations, and address the limitation in the later part}
\label{demographics}
\end{table*}
%They also coded information including demographics (e.g, gender, age and occupation) and mental health states inferred from the posts. Each annotator was assigned 1500 posts, and posts that are irrelevant to loneliness experience were not included in the labeled lonely dataset. 
The {per-category} inter-rater similarity is shown in Figure \ref{fig:inter_rater}.  %(more details in Appendix~\ref{appendix:rater}\leqicomment{this appendix seems to be deleted?}).
%Notably, compared to the context, interpersonal and interaction categories, the duration category has lower inter-rater similarity. 
%\leqiedit{,  suggesting the difficulty of labeling this category}. %suggesting a more ambiguous concept than the others... -> didn't want to say this.
%\todo{Leqi: Add a discussion on how certain category has more agreement and how duration has low similarity. Compared to b, d, a and c are more ambiguous.}
%\todo{Yunfan: Histogram of labels for each category for all posts.}
% \todo{Sherry: coding scheme in the supplementary materials for reviewers}
% \todo{Sherry: Justification of Coding: For R1, ``Perhaps most importantly, because of the central nature of the coding scheme for the novelty of the paper, it would be useful to also unpack why the authors focused on the specific aspects that they propose, and not on others---for example, by explaining why discoursive role and intent of the posts are considered aspects of loneliness. Similarly, the authors could specify what kind of loneliness-related content the codebook is intended for, for example, self-disclosures of loneliness versus requests of support versus discussions around the theme of loneliness.''} \sherry{done}

% \begin{table}[!htbp]
% \centering
% \begin{tabular}{@{}cccc@{}}
% \toprule
% transient & enduring & ambiguous & NA    \\ \midrule
% 22.15\%   & 29.75\%  & 24.54\%   & 23.54\% \\ \bottomrule
% \end{tabular}
% \caption{Label distribution for ``duration"}
% \end{table}

% \begin{table}[!htbp]
% \centering
% \begin{tabular}{@{}ccccc@{}}
% \toprule
% social  & physical & somatic & romantic & NA     \\ \midrule
% 64.83\% & 1.97\%   & 1.86\%  & 25.12\%  & 6.20\% \\ \bottomrule
% \end{tabular}
% \caption{Label distribution for ``context"}
% \end{table}

% \begin{table}[!htbp]
% \centering
% \begin{tabular}{@{}ccccc@{}}
% \toprule
% romantic & friendship & family & peers  & NA      \\ \midrule
% 23.86\%  & 48.57\%    & 4.62\% & 2.45\% & 20.48\% \\ \bottomrule
% \end{tabular}
% \caption{Label distribution for ``interpersonal"}
% \end{table}

% \begin{table}[]
% \centering
% \resizebox{\columnwidth}{!}
% {\begin{tabular}{@{}ccccc@{}}
% \toprule
% seek advice & provide support & seek val. \& aff. & reach out & non-directed \\ \midrule
% 9.89\%      & 3.83\%          & 6.57\%            & 12.26\%   & 67.42\%      \\ \bottomrule
% \end{tabular}}
% \caption{Label distribution for ``interaction"}
% \end{table}

\begin{figure}[ht!]
  \centering
  \begin{subfigure}{\columnwidth}
    \centering
    \begin{tikzpicture}[
      node distance=15mm,
    ]
      \node (a) {A post};
      \node (b) [below left of=a] {Non-Lonely};
      \node (c) [below right of=a] {Lonely};
      \node (d) [below of=c] {Fine-Grained Loneliness Categories};
      \node (e) at ([xshift=-1.1cm, yshift=.1cm]d){};
      \node (f) at ([xshift=-0.2cm, yshift=.1cm]d){};
      \node (g) at ([xshift=1.5cm, yshift=.1cm]d){};
      \node (h) at ([yshift=0.5cm,xshift=0.3cm]d) {$\ldots$};

      \draw (a) to (b)
            (a) to (c)
            (c) to (e)
            (c) to (f)
            (c) to (g)
      ;
    \end{tikzpicture}
    \caption{Hierarchical annotations of a post.}
  \end{subfigure}
  \begin{subfigure}{\columnwidth}
    \centering
      \linespread{0.7}\selectfont\centering
    \begin{tabular}{@{}cc@{}}
      \toprule
      \textbf{Category}          & \textbf{Labels}                              \\ \midrule
      {Duration}      & Transient, Enduring, Ambiguous, NA      \\ \midrule
      {Context}       & Social, Physical, Somatic, Romantic, NA \\ \midrule
      {Interpersonal} & Romantic, Friendship, Family, Peers, NA \\ \midrule
      {Interaction} & \begin{tabular}[c]{@{}c@{}}Seek Advice, Provide Support, \\ Seek Val. \& Aff., Reach Out, Non Directed\end{tabular} \\ \bottomrule
      \end{tabular}
    \caption{Fine-grained loneliness categories and their labels. ``Seek Val. \& Aff.'' denotes ``Seek Validation \& Affirmation". NA indicates not applicable since posts can be irrelevant to such categories or labels.
    %\todo{Add N/A.}\sherry{done}
    } 
  \end{subfigure}
  \caption{The hierarchical labeling structure of a post and its fine-grained loneliness categories. (a) A post is first labeled as lonely or non-lonely. (b) The categories of loneliness will then be annotated with their associated labels.}
  \label{fig:trees}
\end{figure}

% \begin{figure}[ht!]
%     \includegraphics[scale=0.4, center]{figures/inter_rater_cos_similarity.pdf}
%     \caption{Inter-rater similarities measured by the averaged cosine similarities between raters' annotations (more details in Appendix~\ref{appendix:rater}).
%     \todo{Change the similarities to per category similarity. Add a sentence to say the higher, the better. Range between 0 and 1.}}
%     \label{fig:inter_rater}
% \end{figure}

%To perform proof-of-concept inferences using our method proposed in Section \ref{sec:methodology}, we used an unlabeled dataset, which includes all the posts from the final dataset excluding the labeled posts. The unlabeled dataset contains 183,979 posts in total, with 82,037 posts from r/lonely, 3,278 posts from r/loneliness, and 98,664 posts from r/college.

\section{Fine-Grained Loneliness Classification}
\label{sec:methodology}

% In this section, 
We used FIG-Loneliness to 
build models for fine-grained loneliness classification.
%step toward characterizing 
%loneliness by formulating it as fine-grained distributional learning problems. 
Our approach works with two unique features of our dataset:
(1) hierarchical: as explained in Section~\ref{sec:dataset}, and % each post is first classified as lonely or non-lonely, and then only among the lonely data we are to classify them for the fine-grained loneliness labels (Figure \ref{fig:trees});
(2) distributional: instead of being one-hot encoded, the label value for each loneliness categories (i.e., duration, context, interpersonal and interaction) corresponds to a distribution of annotations (more details in Section~\ref{subsec:distributional-labels}).% . For example, the context of a lonely post can be both social and romantic, which can be represented as a distributional label 
%Specifically, we start with the necessities of classifying fine-grained loneliness, elaborate the data processing procedure performed on the labeled dataset, then formulate our problem, and finally present strong baselines to solve it.
% The learned fine-grained loneliness classifiers have strong performance and are then used for loneliness characterization (Section~\ref{sec:characterization}).

\subsection{Distributional Fine-Grained Loneliness Labels}\label{subsec:distributional-labels}

\begin{figure}[ht!]
  \begin{subfigure}{\columnwidth}
    \begin{tcolorbox}[enhanced, drop shadow southwest]
      {\textit{\small ``Have you ever feel bad when your friend talking about her crush? I think I am a introvert but I have being alone. I want someone beside whom I can go out or brag everything. But after I break up with my girlfriend, I feel like no one beside me anymore. And now there's that friends and they are sharing their feeling about their crush. And it make me feel something I don't know. I know I am not in love with them but hearing them talking about someone make me feel hurt. What should I do? Is there something wrong with me?''}}
    \end{tcolorbox}
    \caption{An annotated lonely post.}
  \end{subfigure}
  \begin{subfigure}{\columnwidth}
    \includegraphics[width=\linewidth, center]{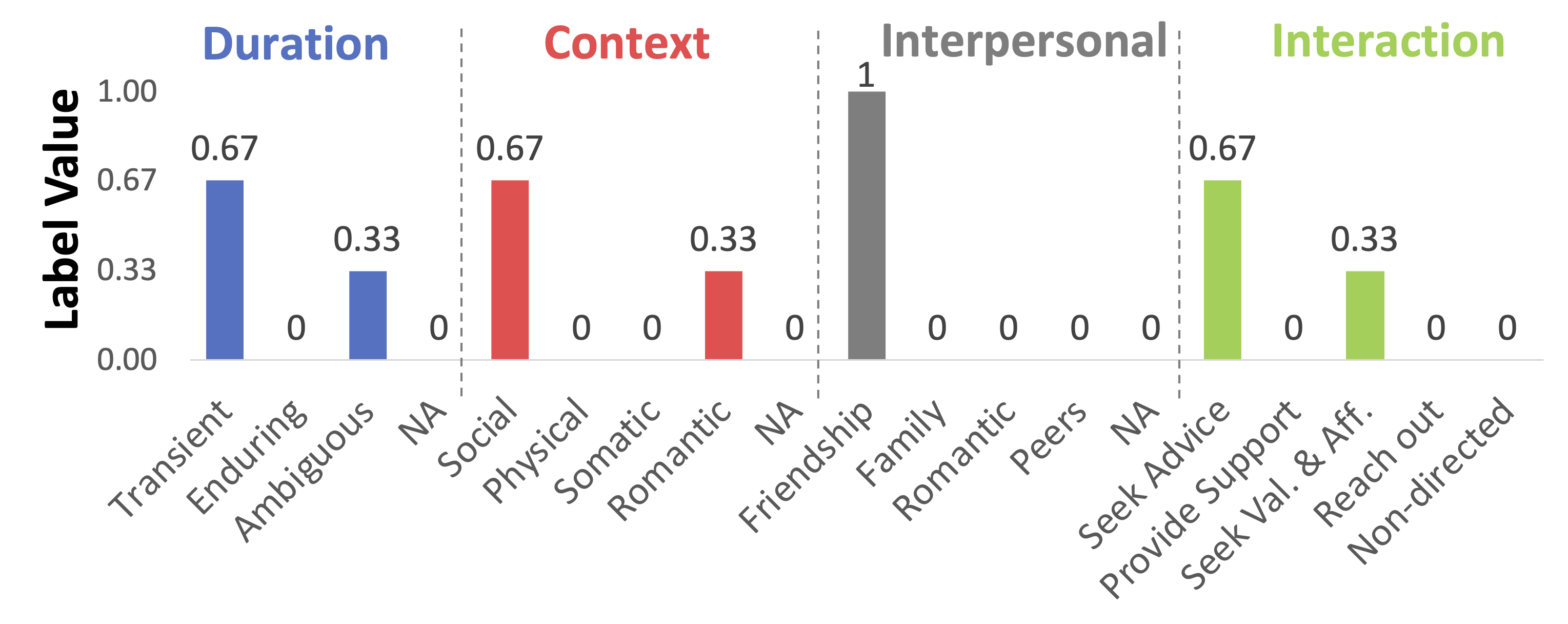}   
    \caption{Distributional label values of the post.
    %\todo{Add N/A.} \sherry{done}
    }
  \end{subfigure}
%   \hfill
%   \begin{subfigure}{0.49\columnwidth}
%     \includegraphics[width=\linewidth, center]{figures/context_distribution_label_illustration.pdf} 
%     \caption{Context}   
%   \end{subfigure}
  \\
%   \begin{subfigure}{0.49\columnwidth}
%     \includegraphics[width=\linewidth, center]{figures/interpersonal_distribution_label_illustration.pdf}
%     \caption{Interpersonal}    
%   \end{subfigure}
%   \hfill
%   \begin{subfigure}{0.49\columnwidth}
%     \includegraphics[width=\linewidth, center]{figures/interaction_distribution_label_illustration.pdf}  
%     \caption{Interaction}  
%   \end{subfigure}
  \caption{ An illustration of a post being annotated by three annotators. (a) The content of an exampled lonely post. (b) Distributional values of the post for all categories. For duration, two raters annotated ``transient'' and one annotated ``ambiguous''. For context, two raters annotated ``social'' and one annotated ``romantic''. For interpersonal relationship, three raters annotated ``Friendship''. For interaction styles, two raters annotated ``seek advice'' and one annotated ``seek validation and affirmation''. NA indicates not applicable.}
  
%   For duration, two raters annotated ``transient'' and one annotated ``ambiguous''. For context, two raters annotated ``social'' and one annotated ``romantic''. For interpersonal relationship, three raters annotated ``Friendship''. For interaction styles, two raters annotated ``seek advice'' and one annotated ``seek validation and affirmation''.
  \label{fig:distribution_label_illustration}
\end{figure}

% As mentioned in Section \ref{sec:dataset}, each Reddit post will first be labeled as lonely or non-lonely. Then, the lonely posts will be 
% coded in four categories described above. 
For each loneliness category, a post might be assigned with different labels by different annotators. % For example, in the context category, 
% a post can be labeled as both 
% social and romantic. Instead of attributing a single value (e.g., using majority voting) to each loneliness category, 
% we used \emph{distribution} over all values in that category. 
% For each category of a post, we normalized the cumulative votes so that the sum of the label vector entries is $1$.
% For example, if a post is annotated by three people, 
% and they believe the contexts to be social, physical and romantic respectively, 
% then the label vector will be $(1/3, 1/3, 0, 1/3)$ since contexts have four possible values (i.e., social, physical, somatic and romantic). 
% Note that these vectors are non-negative and have entries sum up to one, which makes them belong to a probability simplex.
For example, in Figure~\ref{fig:distribution_label_illustration}, the given post was annotated by three annotators. The label value represents the fraction of the annotators who assigned the given label. In this example, the post was coded as ``seeking advice'' by two annotators, and ``seeking validation and affirmation'' by one annotator, which corresponds to $(2/3, 0, 1/3, 0, 0)$ for the label values.
We do so mainly for two reasons.
First, %Humans make decisions with uncertainties. 
a distribution over the labels can be more informative than a single label alone. As shown in~\citet{GeorgeLakoff}, when a unanimous decision is hard to reach, information from different perspectives is important.
% perceptive discrepancies can be morphologically informative, given the difficulties to achieve unanimous decisions among different individuals. 
Second, samples with distributional labels can help train better models. 
Recent studies in natural-image classification, image-based diagnostic, and age estimation have shown that adopting 
distributional labels %for training samples 
results in learning more robust classifiers with better generalization performance, even when using a small amount of labeled instances~\citep{ldl,Peterson2019HumanUM,pmlr-v139-akbari21a}.

\subsection{Hierarchical Distributional Learning}\label{subsec:problem-formulation}

Our problem comprises two learning tasks: a loneliness binary classification task followed by four tasks of distributional fine-grained loneliness classification.
Inspired by recent advances in distributional learning~\citep{ldl} 
and hierarchical learning~\citep{hmcn},
we provide a framework for Hierarchical Distributional-label Learning (HDL). %along with the neural network architecture (HDLN) used for our specific task (Section~\ref{subsec:training}). 
The HDL framework aims at learning the distributions of structured labels. To describe how HDL works under the setup of our problem, we introduce the following notations. 
Let $(\mathbf{x}^{(i)}, \mathbf{\mathcal{P}}^{(i)})$ denote the $i$-th sample 
with $\mathbf{x}^{(i)}$ being the (tokenized) Reddit post 
and $\mathbf{\mathcal{P}}^{(i)}$ being the set of labels for the post. 
The set of labels can be split into two parts $\mathbf{\mathcal{P}}^{(i)} = \{\mathcal{P}_{lonely}^{(i)}, \mathcal{P}_{f.g.}^{(i)}\}$
where $\mathcal{P}_{lonely}^{(i)}$ gives the loneliness distributional label (i.e., lonely or non-lonely)
and $\mathcal{P}_{f.g.}^{(i)}$ provides the fine-grained ones. $\mathcal{P}_{f.g.}^{(i)}$ contains four parts 
where each part $\mathcal{P}_{c}^{(i)}$
corresponds to the distributional label for category $c \in \mathcal{C} = \{\text{duration, context, interpersonal, interaction}\}$. For a non-lonely post, $\mathcal{P}_{c}^{(i)}$ is an all-zero vector; for a lonely post, 
$\mathcal{P}_{c}^{(i)}$ is non-negative, and its entries are summed up to $1$.
Finally, for each sample $\mathbf{x}^{(i)}$, 
its predicted set of labels is denoted to be $\widehat{\mathbf{\mathcal{P}}}^{(i)}$.
Using these notations, the objective of our specific HDL task is to minimize $\mathcal{L} (\mathbf{\mathcal{P}},\widehat{\mathbf{\mathcal{P}}} )$ defined as
\begin{align}\label{eq:HDL-objective}
   \frac{1}{N} \sum_{i=1}^N \big(\ell(\mathcal{P}^{(i)}_{lonely}, \widehat{\mathcal{P}}^{(i)}_{lonely})
   + \frac{1}{|\mathcal{C}|}\sum_{c \in \mathcal{C}}\ell(\mathcal{P}_{c}^{(i)}, \widehat{\mathcal{P}}^{(i)}_{c}) \big),
\end{align}
where $N$ is the number of samples and $\ell$ is a loss function that measures the distance between two distributions (we used cross-entropy loss in our experiment).%.

\subsection{Hierarchical Distributional Learning Methods}\label{subsec:training}

We consider two BERT-based methods for tackling HDL:
a BERT + MLP method 
that adds an MLP classifier on top of a pre-trained BERT model~\citep{bert} for each distributional label;
a Hierarchical Distributional Learning Network (HDLN)
which is an adaption from HMCN-F network \citep{hmcn}. For our baseline model, we use bidirectional LSTM \citep{lstm} with an MLP classifier on top of it to learn each distributional label. 
%\yunfan{DONE} \todo{Add baseline: description and results.}

\begin{figure}
  \includegraphics[width=\columnwidth, center]{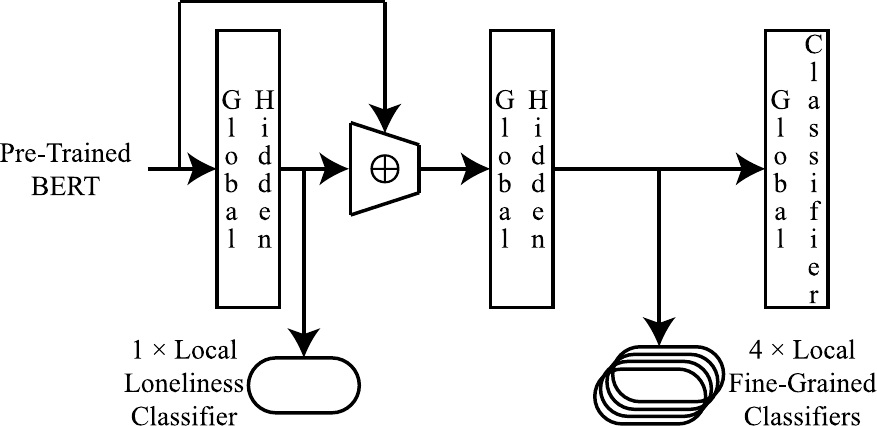}
  \caption{Architecture of the proposed HDLN (more details in Appendix~\ref{appendix:architecture}). $\oplus$ denotes the vector concatenation operation. All local classifiers are two-layer MLPs. ``Global Hidden" represents hidden layers in the global information flow.}
  \label{fig:hdln_arc}
\end{figure}

Both BERT + MLP and HDLN use a pre-trained BERT model fine-tuned on our dataset to obtain the embeddings of Reddit posts. The direct method individually minimizes cross-entropy loss for each classifier. 
On the other hand, HDLN incorporates label hierarchy into its architecture and concurrently outputs multiple distributional predictions.
%Extracted embeddings fed into the HDLN come from a pre-trained BERT model~\citep{bert} fine-tuned on our dataset. 
As illustrated in Figure~\ref{fig:hdln_arc}, HDLN contains five local classifiers and one global classifier. Each local classifier individually predicts one of the five distributions (i.e., the loneliness binary distribution and four fine-grained distributions) and we denote the set of predicted distributional labels from the local classifiers to be  $\widehat{\mathbf{\mathcal{P}}}_{L}$. 
The global classifier predicts the concatenation of all five distributional labels, denoted as   $\widehat{\mathbf{\mathcal{P}}}_{G}$. 
Using both global and local information, HDLN is then trained by minimizing a joint loss function $\frac{1}{2}\mathcal{L} (\mathbf{\mathcal{P}},\widehat{\mathbf{\mathcal{P}}}_L ) + \frac{1}{2}\mathcal{L} (\mathbf{\mathcal{P}},\widehat{\mathbf{\mathcal{P}}}_G )$ where $\mathcal{L}$ is defined in Eq.~\eqref{eq:HDL-objective}. 

% The goal of HDLN is to %\emph{simultaneously} 
% learn multiple classifiers 
% that output distributional labels, rather than to train each classifier independently.
% % As illustrated by the tree structure of our label hierarchy, each post has two levels of labels (Figure~\ref{fig:trees}).
% % The first and second levels contain loneliness binary labels and the labels for four fine-grained loneliness categories, respectively.
% In addition to a better computational efficiency (individually training a neural network for each category is costly),
% HDLN also enables the information of a parent node (i.e., the node containing ``lonely'' and ``non-lonely'' )  being shared among its child nodes (i.e., the nodes representing fine-grained loneliness categories). 

% Finally, 
Following~\citet{hmcn},
once HDLN is trained, we use a linear combination of the outputs from the global and local classifiers as the final predicted distributions $\widehat{\mathcal{P}}_F$, i.e.,
\begin{align}
  \widehat{\mathbf{\mathcal{P}}}_{F} = \beta \widehat{\mathbf{\mathcal{P}}}_{L} + (1-\beta) \widehat{\mathbf{\mathcal{P}}}_{G},
\end{align}
where $\beta \in [0, 1]$ controls the amount of local and global information used in the final prediction.
%\yunfan{better to place this description at the first paragraph of this subsection?} %\todo{Add baseline: description and results.}

% After training, the set of final distributions $\widehat{\mathbf{\mathcal{P}}}_{F}$ used to answer specific questions (e.g., ``does a given post express loneliness?'' and ``what is the composition of the loneliness context?'') can be obtained from
% \begin{align*}
%   \widehat{\mathbf{\mathcal{P}}}_{F} = \beta \widehat{\mathbf{\mathcal{P}}}_{L} + (1-\beta) \widehat{\mathbf{\mathcal{P}}}_{G},
% \end{align*}
% where $\beta \in [0, 1]$ controls the mixture of local and global information \citep*{hmcn}.

\subsubsection{Training}
%Our labeled dataset includes 5,793 valid examples. We randomly split it into a train set (4,055 examples), a dev set (1,158 examples), and a test set (580 examples). We trained models on the train set, fine-tuned performances on the dev set, and report results on the test set.
To train the models, we split FIG-Loneliness into $70\%$ for training, $20\%$ for validation (e.g., hyper-parameter tuning) and $10\%$ for testing. 
We used AdamW \citep{adamw} as our optimizer with a warm-up ratio of $0.1$ and learning rate annealing from $2 \times 10^{-5}$. 
More details of the training procedure are in Appendix~\ref{appendix:training}.
%We adopted a batch size of $16$ and did not use any weight decay. Each training procedure included ten runs with different random seeds.

% We compare HDLN with a baseline model (BERT + MLP) 
% that adds a single-layer MLP classifer on top of the pre-trained BERT model for each distributional label. We minimized cross-entropy loss for each classifer independently. 

\subsubsection{Results}
%\todo{Update Results. Also emphasize HDLN's computational advantage.}
Tables~\ref{tab:eval_lonely} and \ref{tab:eval_fine_grained} show the results on the test set averaged over $3$ random seeds.
%Due to space constraints, 
%we only report results when $\beta=1$
%(for $\beta < 1$, see Appendix~\ref{appendix:results}).
For binary loneliness classification, HDLN
outperformed BERT + MLP and the LSTM baseline in all metrics. To rule out the possibility that the models were only learning the characteristics of different subreddits, we applied HDLN to the test dataset in the college subreddit only, and yielded an F1-score of $93.3\%$
and an accuracy of $99.8\%$.
For fine-grained classification, 
BERT + MLP performed better overall. % but with \emph{much shorter} ($5 \times$ shorter) training time.
% Since an individual BERT + MLP model was required for each distributional label, the equivalent training time of BERT + MLP models were $2.5$ times longer than that of the HDLN model (each individual training of a BERT + MLP model was a half of HDLN's, but we had to train five BERT + MLP models).
% BERT + MLP performed better than HDLN in almost all categories with \emph{much shorter} ($5 \times$ shorter) training time.
However, it is worth noting that a separate BERT + MLP model is trained for each fine-grained category.
Thus, a potential reason for BERT + MLP to outperform HDLN is that each BERT + MLP model has more optimized training parameters specifically for its corresponding fine-grained category. %For example, we terminated each model training right before overfitting occurred, which can occur at different time for different categories. when we individually trained BERT + MLP models, we terminated the training of one model when it exactly converged.
% The total training time of the HDLN model was $2.5$ times shorter than that of the BERT + MLP models (see Appendix~\ref{appendix:training}). However, as 
Despite the lower fine-grained classification accuracy, only a single HDLN model needs to be trained for all categories, 
thus significantly shortening both training and inference time  
(Appendix~\ref{appendix:training}).
In addition, we found that learning the duration category was challenging for all models, consistent with the lower inter-rater similarity observed among human annotators (Figure~\ref{fig:inter-rater-duration}).
% All evaluation metrics are defined in Appendix~\ref{appendix:results}.
{In all subsequent sections, we will use HDLN for binary classification and BERT + MLP for fine-grained loneliness predictions.}

\begin{table}[ht]
  \centering
  \linespread{0.7}\selectfont\centering

\begin{tabular}{@{}ccccc@{}}
\toprule
                                                             & Acc. $\uparrow$                                                        & Precis. $\uparrow$                                                     & Rec. $\uparrow$                                                        & F1 $\uparrow$                                                          \\ \midrule
\begin{tabular}[c]{@{}c@{}}LSTM\\ Baseline\end{tabular}      & \begin{tabular}[c]{@{}c@{}}0.9024\\ $\pm$ 0.0014\end{tabular}          & \begin{tabular}[c]{@{}c@{}}0.8864\\ $\pm$ 0.0030\end{tabular}          & \begin{tabular}[c]{@{}c@{}}0.9014\\ $\pm$ 0.0018\end{tabular}          & \begin{tabular}[c]{@{}c@{}}0.8938\\ $\pm$ 0.0014\end{tabular}          \\
                                                             &                                                                        &                                                                        &                                                                        &                                                                        \\
\begin{tabular}[c]{@{}c@{}}BERT\\ + MLP\end{tabular}         & \begin{tabular}[c]{@{}c@{}}0.9722\\ $\pm$ 0.0046\end{tabular}          & \begin{tabular}[c]{@{}c@{}}0.9538\\ $\pm$ 0.0117\end{tabular}          & \begin{tabular}[c]{@{}c@{}}0.9870\\ $\pm$ 0.0036\end{tabular}          & \begin{tabular}[c]{@{}c@{}}0.9700\\ $\pm$ 0.0048\end{tabular}          \\
                                                             &                                                                        &                                                                        &                                                                        &                                                                        \\
\begin{tabular}[c]{@{}c@{}}HDLN\\\end{tabular} & \textbf{\begin{tabular}[c]{@{}c@{}}0.9763\\ $\pm$ 0.0041\end{tabular}} & \textbf{\begin{tabular}[c]{@{}c@{}}0.9609\\ $\pm$ 0.0045\end{tabular}} & \textbf{\begin{tabular}[c]{@{}c@{}}0.9883\\ $\pm$ 0.0063\end{tabular}} & \textbf{\begin{tabular}[c]{@{}c@{}}0.9744\\ $\pm$ 0.0045\end{tabular}} \\
%                                                              &                                                                        &                                                                        &                                                                        &                                                                        \\
% \begin{tabular}[c]{@{}c@{}}HDLN\\ ($\beta = .5$)\end{tabular} & \begin{tabular}[c]{@{}c@{}}0.9757\\ $\pm$ 0.0046\end{tabular}          & \begin{tabular}[c]{@{}c@{}}0.9597\\ $\pm$ 0.0061\end{tabular}          & \begin{tabular}[c]{@{}c@{}}0.9883\\ $\pm$ 0.0063\end{tabular} & \begin{tabular}[c]{@{}c@{}}0.9738\\ $\pm$ 0.0050\end{tabular}          \\
%                                                               &                                                                        &                                                                        &                                                                        &                                                                        \\
% \begin{tabular}[c]{@{}c@{}}HDLN\\ ($\beta = 1$)\end{tabular}  & \begin{tabular}[c]{@{}c@{}}0.9757\\ $\pm$ 0.0046\end{tabular}          & \begin{tabular}[c]{@{}c@{}}0.9597\\ $\pm$ 0.0061\end{tabular}          & \begin{tabular}[c]{@{}c@{}}0.9883\\ $\pm$ 0.0063\end{tabular} & \begin{tabular}[c]{@{}c@{}}0.9738\\ $\pm$ 0.0050\end{tabular}          \\

\bottomrule
\end{tabular}

  \caption{Results for loneliness binary classification. ``Acc.": Accuracy. ``Precis.": Precision. ``Rec.": Recall. $\uparrow$ indicates higher values are better. 
  }
  \label{tab:eval_lonely}
\end{table}

% \begin{table*}[ht]
%   \centering
%   \begin{tabular}{@{}lllll@{}}
%     \toprule
%                       & Precision $\uparrow$     & Recall $\uparrow$        & F1 $\uparrow$            & Accuracy $\uparrow$      \\ \midrule
%     BERT + MLP         & 0.9753 ± 0.0063          & 0.9644 ± 0.0132          & 0.9697 ± 0.0047          & 0.9720 ± 0.0042          \\
%     HDLN ($\beta$ = 1) & \textbf{0.9759 ± 0.0048} & \textbf{0.9755 ± 0.0060} & \textbf{0.9757 ± 0.0032} & \textbf{0.9774 ± 0.0030} \\
%     \bottomrule
%   \end{tabular}
%   \caption{Results for loneliness binary classification. $\uparrow$ indicates higher values are better.}
%   \label{tab:eval_lonely}
% \end{table*}

\begin{table}[ht]
    \centering
  \linespread{0.7}\selectfont\centering
\begin{tabular}{@{}ccc@{}}
\toprule
                   & Accuracy $\uparrow$          & Clark $\downarrow$           \\ \midrule
\multicolumn{3}{c}{Duration}                                                     \\ \midrule
LSTM Baseline      & 0.4059 $\pm$ 0.0018          & \textbf{1.4491 $\pm$ 0.0009} \\
BERT + MLP         & \textbf{0.5992 $\pm$ 0.0114} & 1.4682 $\pm$ 0.0060          \\
HDLN & 0.4539 $\pm$ 0.0073          & 1.4622 $\pm$ 0.0057          \\ \midrule
\multicolumn{3}{c}{Context}                                                      \\ \midrule
LSTM Baseline      & 0.7198 $\pm$ 0.0000          & 1.9656 $\pm$ 0.0005          \\
BERT + MLP         & \textbf{0.8573 $\pm$ 0.0102} & \textbf{1.9507 $\pm$ 0.0028} \\
HDLN & 0.8560 $\pm$ 0.0220          & 1.9590 $\pm$ 0.0018          \\ \midrule
\multicolumn{3}{c}{Interpersonal}                                                \\ \midrule
LSTM Baseline      & 0.5758 $\pm$ 0.0000          & 1.9369 $\pm$ 0.0010          \\
BERT + MLP         & \textbf{0.7976 $\pm$ 0.0031} & \textbf{1.9077 $\pm$ 0.0009} \\
HDLN & 0.7795 $\pm$ 0.0191          & 1.9123 $\pm$ 0.0015          \\ \midrule
\multicolumn{3}{c}{Interaction}                                                  \\ \midrule
LSTM Baseline      & 0.6459 $\pm$ 0.0000          & 2.0005 $\pm$ 0.0005          \\
BERT + MLP         & \textbf{0.8352 $\pm$ 0.0120} & \textbf{1.9643 $\pm$ 0.0029} \\
HDLN & 0.7237 $\pm$ 0.0138          & 1.9817 $\pm$ 0.0014          \\ \bottomrule
\end{tabular}

    \caption{Results for distributional fine-grained loneliness classification. $\uparrow$ ($\downarrow$) indicates higher (lower) values are better.
    Accuracy and clark are defined formally in Appendix~\ref{appendix:results}.}
    
    %  All distributional metrics used were adopted from \citet{ldl} and defined formally in Appendix~\ref{appendix:results}
    % For a complete table of all results  (e.g., when $\beta > 0$ and on other distributional metrics), see the supplementary materials. 
    %\todo{Update results to newly trained models.}
    %\yunfan{DONE} \todo{Baseline}
    
    \label{tab:eval_fine_grained}
  \end{table}

\section{Loneliness Characterizations}\label{sec:characterization}
Using FIG-Loneliness and the predictive models for fine-grained loneliness classification, we performed quantitative analysis of data from all subreddits (\texttt{r/youngadults}, \texttt{r/college}, \texttt{r/lonely} and \texttt{r/loneliness}) to uncover (1) whether loneliness expressions differ between young adult-oriented communities and communities with diverse age groups; (2) the relationship between users' interaction strategies and forms of loneliness manifested in self-disclosure of loneliness; (3) the impact of COVID-19 on loneliness discourse across the forums.

% (1) the differences of loneliness discourse among these online communities; (2) relationships among the four loneliness categories; (3) the temporal trend of loneliness discourse and the impact of COVID-19 on it.
\label{sec:research-questions}

\subsection{Loneliness Expressions}\label{sec:rq1}
%\subsubsection{RQ1}
% \todo{Tables in this section---Main paper: all data; supplementary: labeled data.} \sherry{I think this can be removed :)}

\begin{figure}[ht!]
  \centering
  \includegraphics[scale=0.5]{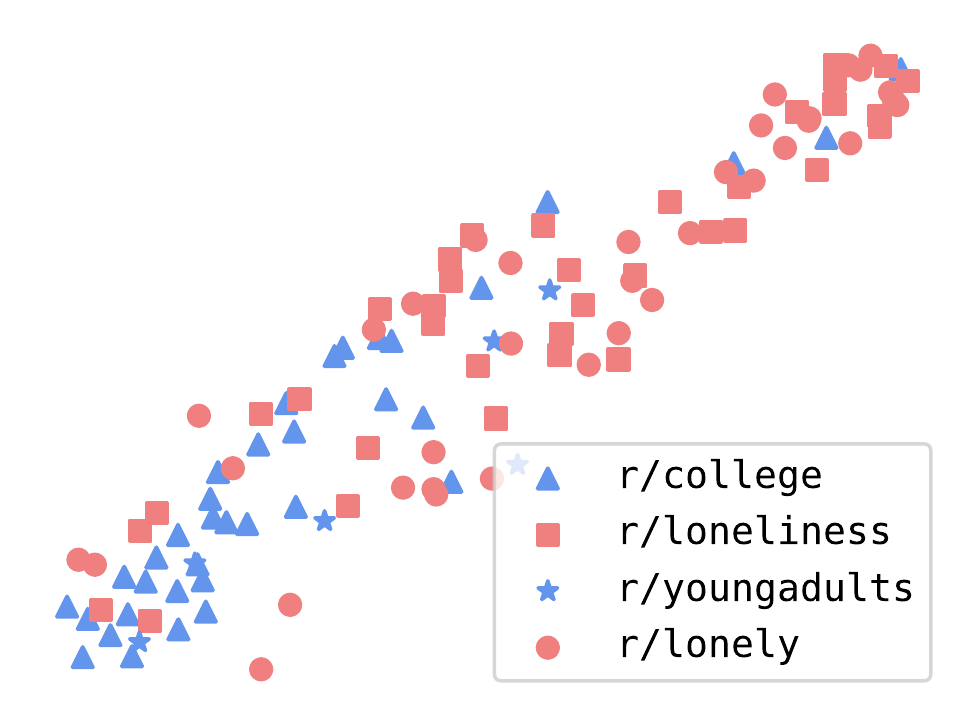}
  \caption{t-SNE embedding of the HDLN model predicted lonely posts across different subreddits. %\yunfan{DONE} \todo{Add r/youngadult.}
  %\todo{try other models. low priority.}
  }
  \label{fig:tsne_subreddit}
\end{figure}

How do young adults focused communities and communities with diverse age groups express loneliness? We explored the differences on loneliness discussions between Reddit communities that consists majority of young adults (\texttt{r/youngadults}, \texttt{r/college}; aged 18-25) and communities of a more diverse age group (\texttt{r/lonely}, \texttt{r/loneliness}). We leveraged the trained HDLN model to visualize the hidden representations of posts from these forums via t-SNE embedding \citep{tsne}. We observed that among all the model predicted lonely posts, the young adult-focused group and the general group have well separated distribution centroids (Figure~\ref{fig:tsne_subreddit}). This suggests that loneliness expressions are more similar within groups than between groups. 
% To gain some understanding of the difference between these two groups of forums, 
% Given that BERT + MLP model performed slightly better than HDLN in fine-grained classification (Section~\ref{subsec:training}),  %for loneliness binary classification to infer the hidden activations before the output layer. 
% Compared with HDLN, BERT + MLP better captures the representational differences between posts from different types of forums. 

To further investigate the sources of such differences, we analyzed the composition of loneliness categories among posts from different subreddits. Table~\ref{tab:constitution_all_category_subreddit} presents the fraction of the posts annotated with each label in a loneliness category for all subreddits. These fractions were obtained by directly averaging the distributional labels of the labeled posts, and were normalized by removing the NA labels. Young adult-focused communities (\texttt{r/college}, \texttt{r/youngadults}) are different from communities of diverse age groups (\texttt{r/lonely}, \texttt{r/lonely}) in all loneliness categories. First, for duration of loneliness, posts from \texttt{r/college} and \texttt{r/youngadults} mentioned more transient loneliness and less enduring loneliness. One explanation is that young adults are more likely to be affected by situational factors, which could become the sources of loneliness \citep{bu2020lonely}. Another explanation is that loneliness related subreddits may attract users with more severe loneliness (see LIWC analyses discussed later in this section). Second, across four loneliness contexts, young adult-oriented communities had more concerns related to somatic and physical loneliness. Interestingly, compared to other forums, college-oriented forum had the highest proportion of loneliness in the social domain and the least proportion of romantic loneliness. This emphasizes the importance of social relationships, especially for the college-age population. Third, regarding the specific relationships mentioned in the posts, we observed that individuals from \texttt{r/youngadults} and \texttt{r/college} experienced more family and peer related loneliness. Finally, we found that those in young adult-focused communities appear to adopt more active coping strategies and problem-focused approach, evidenced by the greater proportions of active coping strategies including seeking advice, validation and reaching out in this group. Moreover, they also received greater amount of support from the communities, indicated by the greater proportion of posts intended to provide support. This may suggest that although young adults are vulnerable to loneliness, they are building resilience in their own communities using the Reddit platform.
% Regarding interaction strategies presented within the loneliness expressions, users from \texttt{r/college} tend to externally request opinions, suggestions, validations, or affirmations. Comparatively, users from other two loneliness-specific subreddits were more likely to act in a non-directed interaction manner (no desire for interacting with other users).
% friendship received the greatest proportion of mentions, which implies that friendship is an important constituent in social relationships. 

We also examined loneliness-associated language markers using the Linguistic Inquiry and Word Count (LIWC) dictionaries \citep{pennebaker2015development}. We individually tested each dictionary output as a predictor for a logistic regression model controlling for year and subreddit, and corrected the standardized regression coefficients using Benjamini–Hochberg adjustment. Across all communities, posts with lonely labels are more present focused (example: \emph{today}, \emph{is}, \emph{now}; $\beta$ = .11; $P <$ .001) and contain more words related to negative emotions (example: \emph{hurt}, \emph{ugly}, \emph{nasty}; $\beta$ = .21; $P <$ .001), sadness (example: \emph{crying}, \emph{grief}, \emph{sad}; $\beta$ = .26; $P <$ .001) and feeling (example: \emph{feel}, \emph{touch}; $\beta$ = .15; $P <$ .001). We also observed that lonely posts are associated with the use of more first-person pronouns ($\beta$ = .18; $P <$ .001) and references to cognitive processes (example: \emph{cause}, \emph{know}, \emph{ought}; $\beta$ = .13; $P <$ .001). Analyses on a group level revealed that the young adult oriented group (\texttt{r/youngadults}, \texttt{r/college}) used more language related to peers such as buddy, coworker and neighbour ($\beta$ = .18; $P <$ .001) and sadness ($\beta$ = .29; $P <$ .001) compared to the other group (peer: $\beta$ = .07; $P <$ .001; sadness: $\beta$ = .23; $P <$ .001 ) in loneliness discourse. This is consistent with the higher proportion observed in \emph{peers} and \emph{transient} related loneliness in the young adult-focused forums.

% \todo{Add Language Association. General public vs young adult.} \sherry{done}

% \begin{table}%[ht!]
%   \centering
%   \setlength{\tabcolsep}{0.5pt}
% \begin{tabular}{@{}ccccc@{}}

% \toprule
%                     & r/youngadults \& r/college & r/lonely \& r/loneliness  \\
%                     \midrule

% \multicolumn{3}{c}{Duration} \\
%  \midrule
% transient           & 30.0 & 29.3\\
% enduring            & 34.2 & 34.5 \\
% ambiguous           & 35.8 & 36.2  \\
%  \midrule
% \multicolumn{3}{c}{Context} \\
%  \midrule
% social              & 77 & 71.6     \\ 
% physical            & 4.7 & 2.3   \\ 
% somatic             & 5.2 & 2.9    \\ 
% romantic            & 13.1 & 23.2    \\ 
%  \midrule
% \multicolumn{3}{c}{Interpersonal} \\
% \midrule
% romantic            & 14.7 & 26.0    \\ 
% friendship          & 63.1 & 63.3    \\ 
% family              & 14.5 & 7.3    \\ 
% peers               & 7.7 & 3.4    \\ 
%  \midrule
% \multicolumn{3}{c}{Interaction} \\
% \midrule
% seek advice         & 17.7 & 10.0                                    \\
% provide support     & 9.7 & 5.1                                   \\
% seek val. \& aff.   & 11.9 & 6.7                                    \\
% reach out           & 18.8 & 16.8                                    \\
% non directed        & 41.9 & 61.4                                \\
% \midrule
% \end{tabular} 
% \caption{Fine-grained category composition. Percentages of all label values within a category are shown. Note that label values were normalized after removing NAs. }
%   \label{tab:constitution_all_category_subreddit}
% \end{table}

\begin{table}%[ht!]
  \centering
  \setlength{\tabcolsep}{3pt}
  \linespread{0.5}\selectfont\centering

\begin{tabular}{@{}ccccc@{}}
\toprule
                    & r/youngadults & r/college & r/lonely & r/loneliness \\
                    \midrule

\multicolumn{5}{c}{Duration} \\
 \midrule
transient           & 31.7 & 29.8 & 29.4 & 28.8  \\
enduring            & 29.5 & 34.8 & 34.5 & 35.2   \\
ambiguous           & 38.8 & 35.4 & 36.2 & 36.0    \\
 \midrule
\multicolumn{5}{c}{Context} \\
 \midrule
social              & 69.8 & 77.9 & 71.6 & 73.1     \\ 
physical            & 3.7 & 4.9 & 2.3 & 3.0    \\ 
somatic             & 4.8 & 5.2 & 2.8 & 3.6    \\ 
romantic    & 21.7 & 12.0 & 23.3 & 20.3    \\ 
 \midrule
\multicolumn{5}{c}{Interpersonal} \\
\midrule
romantic         & 24.3 & 13.5 & 26.1 & 23.1     \\ 
friendship          & 60.2 & 63.5 & 63.3 & 63.5    \\ 
family              & 10.1 & 15.1 & 7.2 & 9.3    \\ 
peers               & 5.4 & 8.0 & 3.4 & 4.2    \\ 
 \midrule
\multicolumn{5}{c}{Interaction} \\
\midrule
seek advice         & 14.4 & 18.1 & 10.0 & 11.1                                      \\
provide support     & 9.2 & 9.7 & 5.0 & 5.9                                    \\
seek val. \& aff.   & 9.4 & 12.2 & 6.6 & 7.5                                    \\
reach out           & 23.8 & 18.2 & 16.9 & 15.5                                    \\
non directed        & 43.2 & 41.8 & 61.5 & 60.0                                    \\
\bottomrule

\end{tabular} 
\caption{Fine-grained category composition. Percentages of all label values within a category are shown. Note that label values were normalized after removing NAs. }
  \label{tab:constitution_all_category_subreddit}
\end{table}

\subsection{Coping Strategies}\label{sec:rq2}
%\subsubsection{RQ2}
How do different forms of loneliness associated with the types of coping strategies? To answer this question, we examined the conditional dependence between authors' interaction intents and the forms of loneliness in the domains of duration, context and interpersonal relationships in FIG-Loneliness. % leveraged the labeled dataset and examined the composition of one category given another category of fine-grained loneliness. 
For example, to study the relationship between interaction strategies and duration of loneliness, for posts with the same duration label, we computed the percentage of them annotated with different interaction labels. In this paper, we focused on two types of coping strategies. Active coping strategies are used to define and resolve the source of stress, and thus include ``reaching out'', ``seeking advice'' and ``seeking validation'' in the interaction labels. Passive coping strategies involve attempts to manage emotional stress without confronting the source of stress, which corresponds to our ``non-directed interaction'' label.

\subsubsection{Duration of Loneliness}

Overall, more than $67\%$ posts did not explicitly seek interactions with other users. Posts labeled as transient loneliness have a greater proportion ($5.3\%$ greater) than enduring loneliness in the use of active strategies. The most used active strategy is ``reaching out'' for transient loneliness ($14.4\%$), and ``seeking advice'' for enduring loneliness ($10.7\%$). A closer examination of the difference in coping strategies between the two types of loneliness revealed that those individuals with transient loneliness make relatively more efforts in reaching out or connecting with others ($6.7\%$ more), and fewer attempts at seeking validation or affirmation from people ($1.6\%$ less). Moreover, posts with enduring loneliness have the greatest proportion of ``non-directed interaction'', indicating a greater use in the passive coping strategies.

% \sherry{add interpretation} This corresponds to  \citet{cacioppo2008loneliness}'s the motivational aspects of loneliness for transient loneliness. According to this view, loneliness is an adaptive trait of a complex social species such as our own that aids in consolidating and reconstructing relationships, which can promote our genetic survival. In transient loneliness, reaching out or seeking for social connection is an advantageous strategy for individuals to improve social relationships.

\subsubsection{Loneliness Context}

Across the four contexts, more than $66\%$ of the posts used passive coping strategies (i.e., non-directed interaction). Seeking advice is the most popular strategy in physical ($12.8\%$) and romantic loneliness ($11.6\%$). In addition, the most used strategies for somatic and social loneliness are seeking validation ($10.6\%$) and reaching out ($14.4\%$).

% Among the interactive posts, seeking advice is the most popular interaction intent, followed by reaching out. Interestingly, of all posts annotated as somatic loneliness, there were more posts seeking validation and affirmation than seeking advice. This suggests that those who experience loneliness due to physical symptoms or conditions are more likely to ask for emotional support rather than seeking advice or connections with others. Moreover, in posts that provide support to deal with loneliness, $85.0\%$ are about social loneliness and $14.0\%$ are about romantic loneliness. This indicates that social loneliness receives a greater level of support for resilience building in online communities. This is consistent with our finding that $82.0\%$ of the posts that were marked as "reaching out" were also labeled as "social loneliness', suggesting that users in the forums are actively responding to others' loneliness concerns.

\subsubsection{Social Relationships for Loneliness}
Regarding interpersonal relationships involved in the loneliness discourse, consistent with the previous findings, majority of the posts used passive coping strategies (over $65\%$). In the posts seeking for interactions, seeking advice is the most common user interaction strategy pertaining to romantic ($11.8\%$), family ($11.0\%$) and peers ($18.6\%$) related relationship issues, whereas reaching out is the most used strategy to cope with loneliness issues raised by friendships ($15.3\%$).

\subsection{Effects of the COVID-19 Pandemic}\label{sec:rq3}
%\subsubsection{RQ3}

% \leqi{(1) Let's first elaborate on this problem a little bit, say that we care about the trend before and after the lockdown and why. 
% (2) Introduce the interrupted time series model: write down the linear model we have used (and acknowledge the strong assumption of this model which I could add)
% and what can be drawn from the model fitting. 
% (3) Provide more information on what data has been used in to obtain our results and report the results (coefficients along with 95\% CI and two plots). Each main conclusion we obtain should be very easy for the reader to catch even if they only read the beginning and ending of each paragraph.}

% \sherry{I agree with Leqi. I can add (1) and write the interpretations of results 
% }

% Previous research found that the negativity within loneliness expressions fluctuates throughout week, showing a natural pattern of sentiment shift across time \citep{mahoney2019feeling}. 
During the COVID-19 pandemic,
understanding changes in mental health needs over time, especially related to loneliness, has become an important and urgent topic. 
The uncertainty in the pandemic, caused by the health threat from the virus and financial stress, has negatively affected many individuals' mental and physical health. 
Moreover, due to the social distancing policy, the prolonged state of physical isolation from peers, teachers and community networks may lead to a significant increase in the feelings of loneliness, especially among young adults \citep{bu2020lonely}. Thus, a better understanding of the trends of different loneliness categories before and during the pandemic can facilitate early intervention and design for reducing loneliness in public. In this section, we investigated the impact of COVID-19 on people's loneliness experiences on Reddit.

\begin{figure}[ht!]
    \begin{subfigure}{0.4\columnwidth}
        \includegraphics[width=\textwidth]{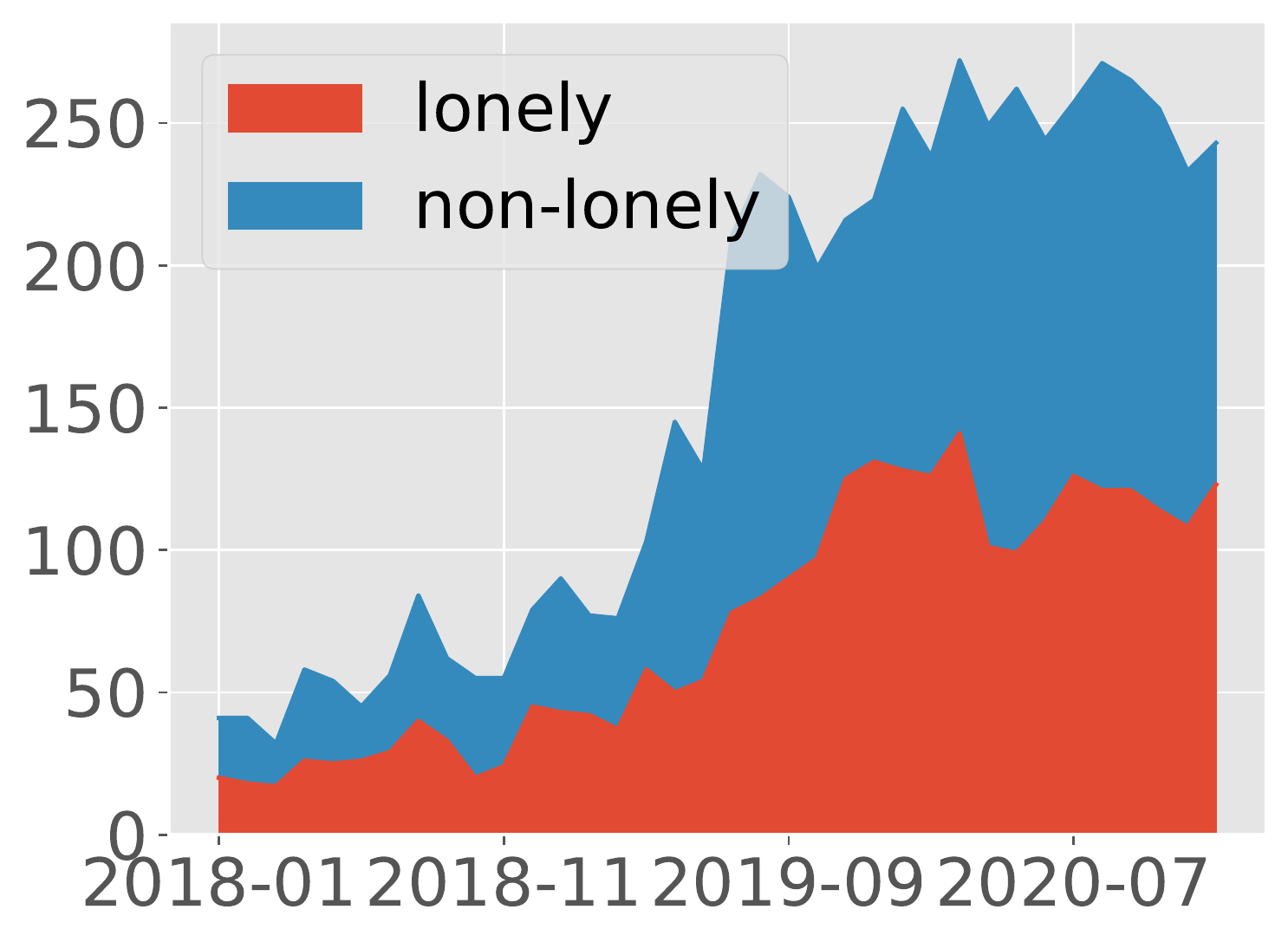}
        \caption{FIG-Loneliness.}
        \label{subfig:stacked_area_labeled}
    \end{subfigure}
    \hfill
    \begin{subfigure}{0.4\columnwidth}
        \includegraphics[width=\textwidth]{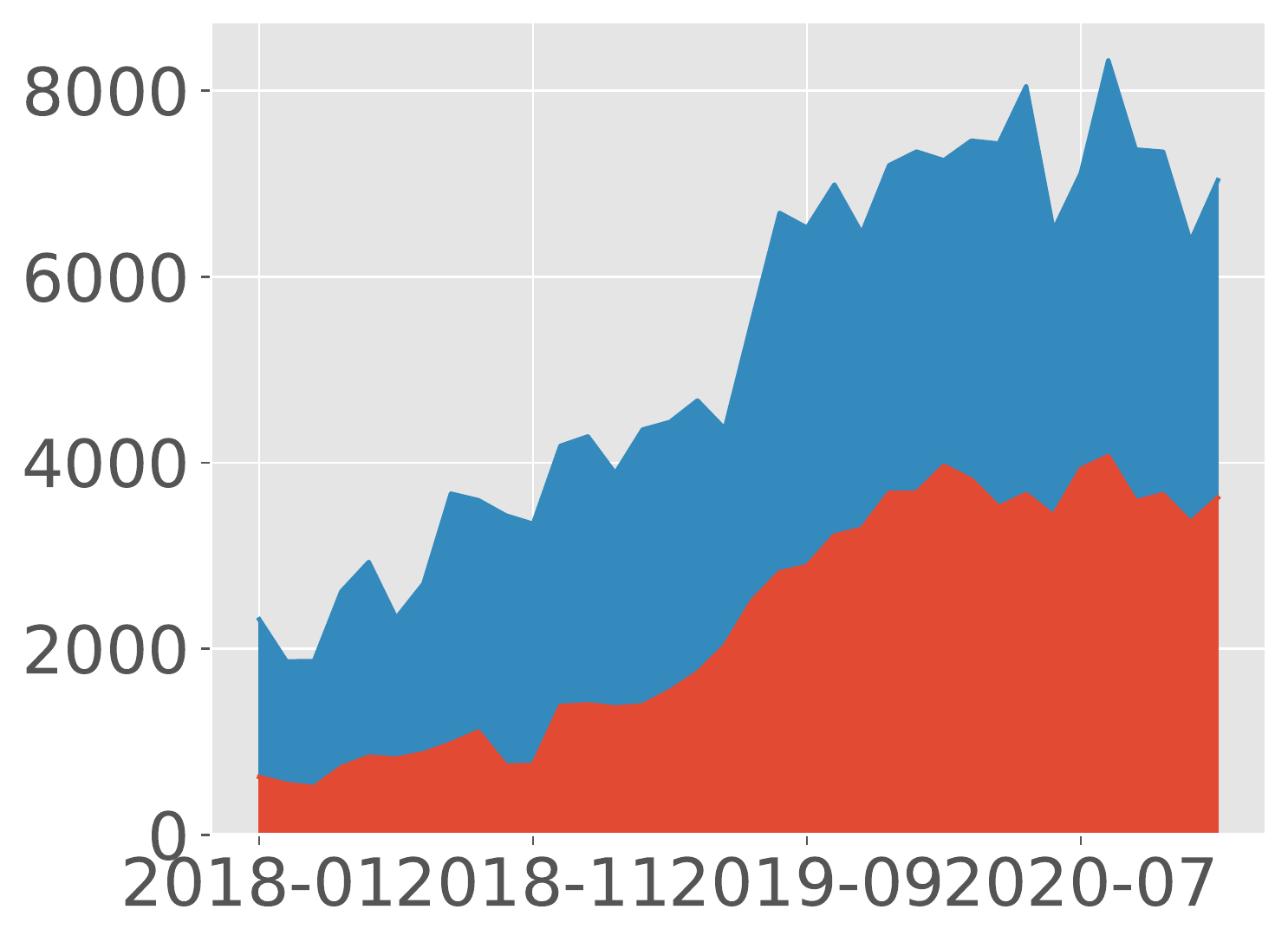}
        \caption{Unannotated data.}
        \label{subfig:stacked_area_unlabeled}
    \end{subfigure}
  \caption{Number of lonely and non-lonely posts over time. (a) Representation of lonely vs. non-lonely posts in FIG-Loneliness. (b) Representation of lonely vs. non-lonely posts for unannotated data using the model predictions from HDLN ($\beta=0$). %\todo{update the figure with new data.\yunfan{done with corrected model}}
  }
  \label{fig:stacked_area}
\end{figure}

To do this, we first examined the temporal trend of loneliness-related discussions on Reddit. We observed that both the total number of posts and loneliness-related discussions in FIG-Loneliness (i.e., the annotated posts) have been growing across \texttt{r/college},  \texttt{r/youngadults}, \texttt{r/lonely} and \texttt{r/loneliness} from January 2018 to December 2019 (Figure \ref{subfig:stacked_area_labeled}). 
A similar pattern was observed in the HDLN model ($\beta = 0$) predicted labels on posts that were not annotated (Figure~\ref{subfig:stacked_area_unlabeled}). %for loneliness 
%For the unlabeled posts, we used the trained HDLN model with $\beta = 1$ to infer the loneliness binary labels (i.e., lonely or non-lonely), and found similar pattern. 
Moreover, the consistent high volume of posts in 2020 suggests that %Reddit has been a popular platform for 
many individuals kept sharing and processing (loneliness) experiences during the pandemic on Reddit. 

Next, we looked at the composition for the forms of loneliness across years in FIG-Loneliness.
%(see Tables , \ref{tab:constitution_pri_ctx_year}, \ref{tab:constitution_pri_intrp_year}, and \ref{tab:constitution_interaction_year}).  
We found some noticeable changes in the types of loneliness people discussed in the pandemic (year 2020) compared to before (year 2018, 2019). Specifically, during the pandemic, there were increases in loneliness expressions on transient loneliness (by $8\%$), and concerns around romantic (context: $5\%$; interpersonal: $6\%$) and family relationship (by $3\%$). This indicates that romantic and family loneliness rises in 2020, perhaps as a result of the quarantine with romantic partners and family in the pandemic. 
% \todo{Need two sets of analyses:
% all subreddit; college and young adult.
% Then, we pick the results that are significant.
% }

To investigate \emph{the immediate effects} of the outbreak of pandemic (the lockdown orders) on different forms of loneliness expressions, using FIG-Loneliness and HDLN model predictions, we 
performed interrupted time-series analysis~\citep{interrupted_time_series_analysis} on the proportions of posts belonging to different category labels across months. %of the posts corresponding to different types of loneliness expressions across months
We took March, 2020 to be the time when the intervention applied
since 
the lockdown and stay-at-home orders were issued in the U.S. then. 
We used the following linear model to fit the data:
% \footnote{We note that using a linear model requires strong assumptions on the data, e.g., the error terms $e$ among the observations are uncorrelated~\citep{mcdowall2019interrupted}.}:
% In this analysis, using the labeled dataset and ``March,  month\footnote{March, 2020 was when the lockdown and stay-at-home orders were issued in the U.S.} by fitting the following regression model
\begin{align*}
  Y = b_0 + b_1 T + b_2 D + b_3 M + e,
\end{align*}
where $Y$ represents the proportions of different types of posts over time,
$T$ represents months since January 2018, 
$D$ is a binary variable indicating whether the intervention is applied (i.e., before or after March, 2020), 
and $M$ is the months passed since the intervention. 
Coefficients $b_0$, $b_1$, $b_2$, and $b_3$ correspond to the initial proportion, the pre-intervention trend, immediate effect of the intervention, and the difference between pre- and post-trends, respectively. $e$ represents an error term.

% ITS analysis statistically investigates the effects of an intervention (in our case, social distancing resulted from the pandemic) on an outcome over time (in our case, the longitudinal proportion of a certain type of loneliness) by establishing and fitting a linear model
% \begin{align*}
%   Y = b_0 + b_1 T + b_2 D + b_3 P + e,
% \end{align*}
% where $Y$ represents the longitudinal data, $T$ a continuous variable indicating time, $D$ a binary variable indicating the outcome collected before or after the intervention, and $P$ a continuous variable indicating time passed since the intervention. Coefficients $b_0$, $b_1$, $b_2$, and $b_3$ are a constant bias, the pre-intervention slope, immediate effect of the intervention, and the difference between pre and post slopes, respectively. $e$ represents an error term.

% \yunfan{brief results from time series analysis and introduction of the method/model}
% We found that the pandemic has \emph{significant} immediate effects on the duration and contexts of loneliness. It also continuously affects the contexts of loneliness expressed from college students. We drew these conclusions by performing the interrupted time series (ITS) analysis \citep{interrupted_time_series_analysis}.

\begin{figure}[ht!]
  \begin{subfigure}{0.4\columnwidth}
    \includegraphics[width=\textwidth, center]{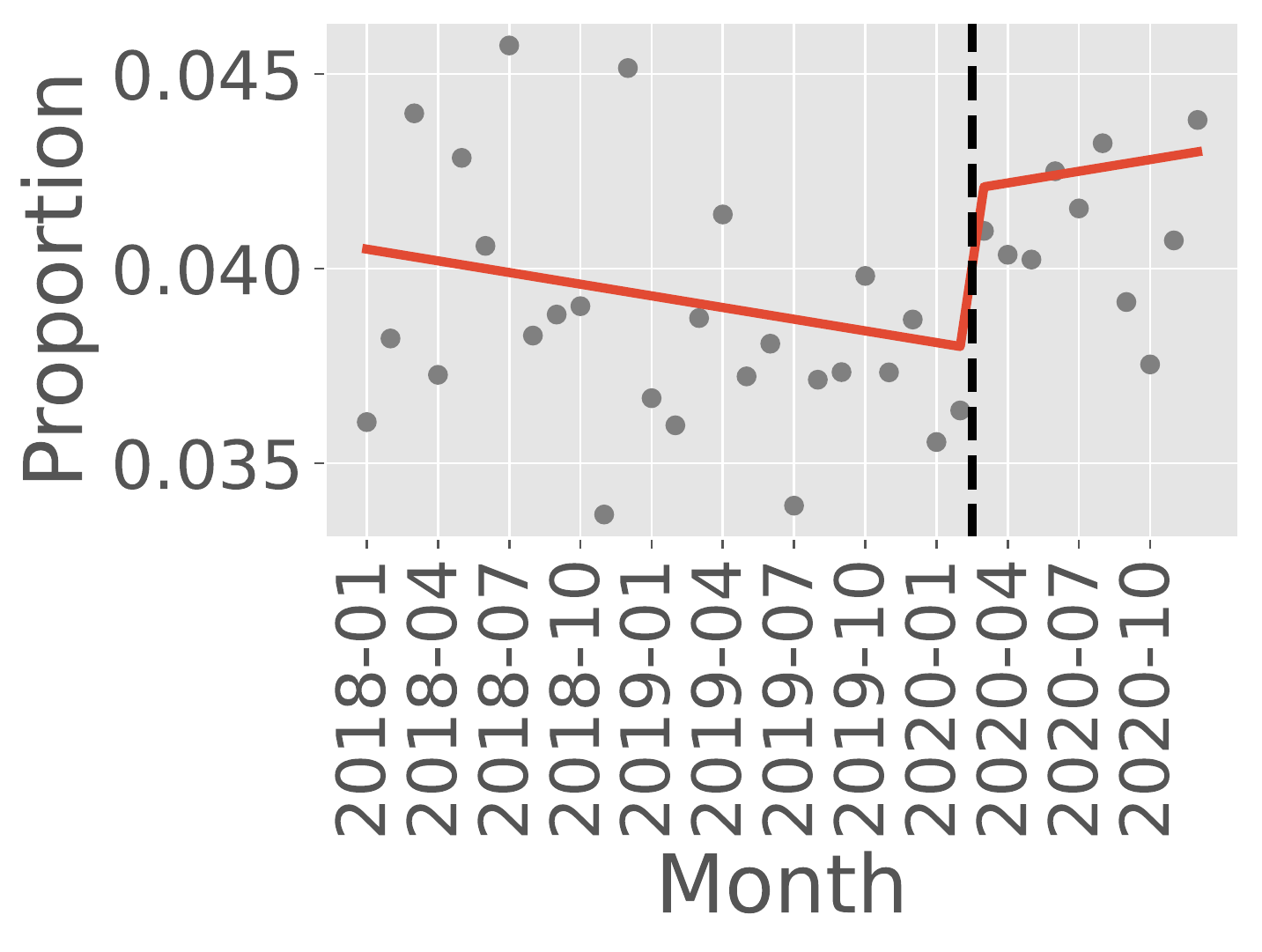}
    \caption{Physical context.} 
    %from \texttt{r/youngadults} and \texttt{r/college}.}
    \label{fig:its_enduring}
  \end{subfigure}
  \hfill
  \begin{subfigure}{0.4\columnwidth}
    \includegraphics[width=\textwidth, center]{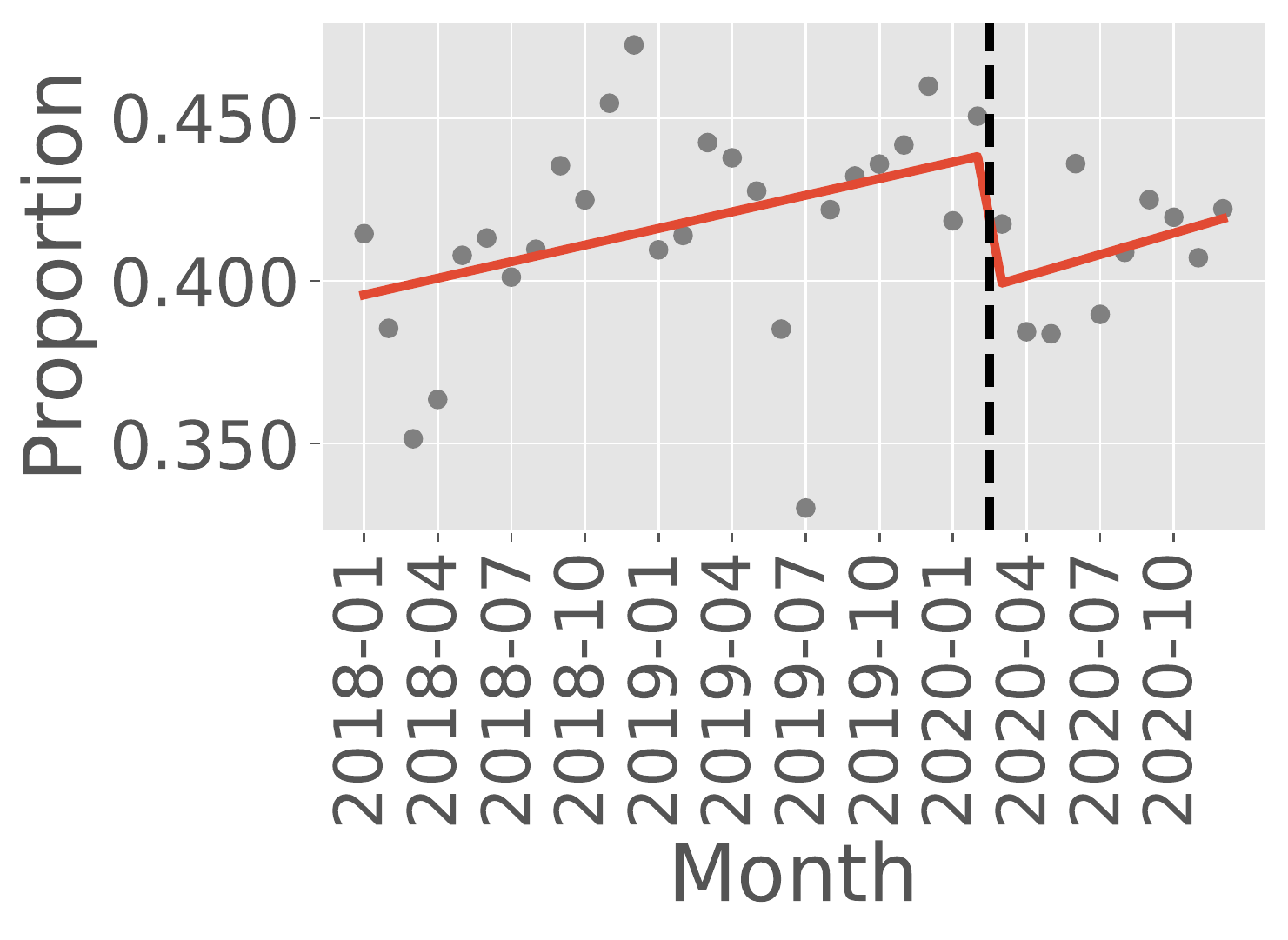}
    \caption{Non-directed interaction.}
    %from \texttt{r/youngadults} and \texttt{r/college}.}
    \label{fig:its_physical}
  \end{subfigure}
  \caption{Trends of fine-grained loneliness category labels for (a) physical context and (b) non-directed interaction in young adult focused forums. Dashed lines represent the intervention (i.e., the national lockdown). 
%   The analysis is specific to \texttt{r/youngadults} and \texttt{r/college}.
%   \todo{
%   1. Update these two figures: enduring + physical context: general v.s. young adult.
%   [low priority 2. Active interaction (seeking advice + reaching out + seeking affirmation)]
%   }
%   \todo{Add caption for dotted line.}
%   \todo{
%   - Duration look at enduring, transient; 
%   - Context look at all labels;
%   - Interaction: Nondirected interaction (passive); 
%   seeking advice + reaching out + seeking affirmation (active)}
%   }
% \yunfan{DONE}
    }
  \label{fig:time_series_analysis}
\end{figure}

Among the four loneliness categories across subreddits, 
context and interaction yielded statistically significant immediate effects, only in the young adult forums (\texttt{r/youngadults} and \texttt{r/college}). We speculate that the COVID-19 school lockdown may lead to increased loneliness in young adults such as college students, as a result of changes in physical environments. Specifically, the proportion of \emph{physical} loneliness had a sudden increase from $3.8\%$ to $4.2\%$ in March, 2020 (Figure~\ref{fig:its_enduring}; $95\%$ CI $[-.0004, .0084]$, $P < 0.1$). This suggests that young adults were potentially immediately affected by the lockdown orders and changes in physical environments. In addition, we observed a sudden decrease in the proportion of loneliness expressions related to \emph{non-directed interaction} or passive coping strategies, from $44.0\%$ to $39.9\%$ (Figure~\ref{fig:its_physical}; $95\%$ CI $[-.0855, .0033]$, $P < 0.1$). This could result from an active coping approach adopted by the young adult population, given that they are overall more active in their interaction styles pertaining to loneliness discussions (see Table~\ref{tab:constitution_all_category_subreddit}).

\section{Discussion and Conclusion}

In this study, we provide a large human coded dataset FIG-Loneliness that includes annotations for fine-grained categories of loneliness. Using methods for hierarchical distributional learning, we built multi-label classifiers to examine how different forms of loneliness and users' interaction styles (including coping strategies) manifest in online discourse from two loneliness specific forums (\texttt{r/lonely}, \texttt{r/loneliness}) and two young adults focused forums (\texttt{r/college}, \texttt{r/youngadults}). With FIG-Loneliness and model predictions, we showed the differences in loneliness discourse across these Reddit forums, the relationships between loneliness forums and authors' coping strategies, and the impacts of the pandemic on different loneliness types. 

We found that from 2018 to 2020, Reddit has become an increasingly popular platform for individuals to disclose sensitive information such as personal loneliness experiences. Inferred from loneliness discourse in \texttt{r/college} and \texttt{r/youngadults}, young adults, the primary users of social media, are more likely to use the Reddit platform to seek active interactions with others to cope with loneliness. They also receive some social support which could build resilience in the communities. Compared to other groups, young adults are more likely to experience loneliness in the physical and somatic domains, which highlights the potential harmful effects of loneliness on physical health due to  geographical isolation during the pandemic. Further, family and peer related loneliness appear more prevalent in the young adult group, which can be explored in future research.

Consistent with prior literature, loneliness is associated with the use of first-person pronouns, references to words indicating thinking and reasoning (LIWC cognitive processes dictionary), and language related to negative emotions and sadness in the LIWC dictionaries \citep{guntuku2019studying}. This suggests a preoccupation of self and a potential association between rumination and loneliness, which is also associated with depression. We found that those in young adult-focused forums are more likely to mention words related to \emph{peers} in the loneliness discourse, suggesting that peer support is especially critical in promoting young adults' psychological well-being.

In addition, individuals with different forms of loneliness have different interaction styles. For example, in posts that request explicit social interactions, individuals with transient loneliness are more likely to seek social connections whereas those with enduring loneliness are more likely to seek advice or validation from the communities. Physical and romantic loneliness are also more associated with requests for advice, and somatic loneliness is more associated with validation seeking. These interaction styles are associated with active coping strategies because they are problem-focused. However, the majority of the posts are using an emotion-focused or passive coping approach (no explicit intent for social interactions). Future research efforts should work on methods to help individuals to adopt active coping strategies to alleviate the feeling of loneliness. Those with enduring loneliness can especially benefit from such work as they bear more negative consequences from loneliness, and have a relatively greater use of passive coping strategies. 

Moreover, COVID-19's social distancing and school lockdown orders may have a greater immediate effect on individuals from \texttt{r/college} and \texttt{r/youngadults}, as evidenced by a sudden increase in loneliness related to physical context and requests for active interactions within the communities. Linking our observation that those from young adult-focused forums experience relatively more physical loneliness, we speculate that young adults may be more sensitive to the influence of geographical isolation (e.g., school lockdown or community changes). Future studies should examine the risk factors involved in changes of physical environments for young adults' loneliness.

It is worth emphasizing that loneliness is a multidimensional concept and can be measured in different ways. The distinction between various forms of loneliness has been emphasized in psychological literature as it bears on understanding unique causes, manifestations, and consequences of each form. Our study confirms the possibility of automatic detection of different forms of loneliness and coping strategies, and provides a model framework for loneliness screening. The prediction of such information can help identify who may possibly need additional resources, and employ effective early intervention programs for loneliness. In future work, it is important to examine the causal relationship between the specific loneliness forms and their long-term adverse effects. For example, previous literature suggests that individuals who suffer from chronic loneliness exhibit long-term difficulties in interpersonal relationships, relative to transient loneliness which relates to an adaptation to situational changes \citep{de1982types}. This is because certain forms of loneliness, such as transient loneliness, are typically easier to cope with and recover from. It is possible that those who are with certain labels in their loneliness expressions such as ``chronic" and ``somatic" loneliness, are more likely to develop negative psychological outcomes in the long run compared to those with labels including ``transient" and ``physical" loneliness. Similarly, future work may follow-up on the effects of different coping strategies (e.g., seeking advice or validation and reaching out) on loneliness reduction because previous work suggests that coping strategies can make a difference in the psychological outcomes \citep{deckx2018systematic}.

% As suggested in literature, different styles of coping are good predictors of future outcomes \citep{folkman2011oxford}. 
\paragraph{Limitation and Data Disclaimer} 
Several limitations to our work should be noted. We drew loneliness discourse from \texttt{r/college} and \texttt{r/youngadults} as a proxy for studying young adults' loneliness expressions. However, these data may not be representative of young adults' expressions. For example, $16.7\%$ of \texttt{r/college} and \texttt{r/youngadults} posts were written by users below age $18$. Another limitation of our data is that our non-lonely samples are only from \texttt{r/college} and \texttt{r/youngadults}. To capture a broader range of expressions not related to loneliness, one may sample non-lonely posts from a more diverse category of subreddits. 
We also acknowledge that our collected data is from a particular social media platform (Reddit)
and written in English only.
Thus, it may not represent loneliness discourse outside this scope, and may contain cultural, platform-specific biases among many others. 
Since our models are trained on our choice of data, 
the pattern it learned may only be reflective to this particular source of data. The models' ability to generalize to other data sources needs to be further studied.

\paragraph{Acknowledgments} 
We thank Terresa Eun and Weier Wan for their helpful discussions and comments. This work was supported by an unrestricted gift from the nonprofit HopeLab Foundation.
LL is generously supported by an Open Philanthropy AI Fellowship.

\appendix
%\onecolumn
\section{Appendix}

% \subsection{Inter-rater Similarity}
% \label{appendix:rater}

% Inter-rater similarities between any two raters were the average Cosine similarities on their overlapping annotated lonely posts.
% We use inter-rater similarity to examine the similarities in annotations given by any two raters. For every post labeled by both raters, we computed the cosine similarity between their annotations. To obtain inter-rater similarity between the two raters, we averaged cosine similarities over all overlapping posts.
% Concretely, we first found lonely posts annotated by two raters in common. The annotation per post per rater was then one-hot encoded into a vector. Subsequently, for all common posts each with two annotation vectors, Cosine similarity values were computed and then averaged to get the aggregated similarity.

\subsection{Model Architectures}
\label{appendix:architecture}

Architecture hyperparameters we used are given in Table~\ref{tab:arc_hyperparms}.
% for BERT + MLP models (table \ref{tab:arc_hyperparam_mlp}) and the HDLN model (table \ref{tab:arc_hyperparam_hdln}) are listed as below.
BERT + MLP and HDLN models take input from pre-trained BERT models implemented by \citet{huggingface}. We used the \texttt{bert-base-cased} variant with default configurations. The output size of each classifiers equals to the number of supports of corresponding distribution.

\begin{table}[ht!]
\centering
  \linespread{0.7}\selectfont\centering

\begin{tabular}{@{}cc@{}}
\toprule
Hyperparameter          & Value   \\ \midrule
\multicolumn{2}{c}{LSTM Baseline} \\ \midrule
Input Size              & 768     \\
Hidden Size             & 128     \\
\# LSTM Layers          & 1       \\
\# Classifier Layers    & 2       \\ \midrule
\multicolumn{2}{c}{BERT + MLP}    \\ \midrule
Input Size              & 768     \\
Hidden Size             & 50      \\
\# Layers               & 2       \\ \midrule
\multicolumn{2}{c}{HDLN}          \\ \midrule
Input Size              & 768     \\
Global Hidden Size      & 64      \\
Local Hidden Size       & 64      \\
\# Layers               & 2       \\ \bottomrule
\end{tabular}

\caption{Architecture hyperparameters for LSTM Baseline, BERT + MLP, and HDLN models.}
\label{tab:arc_hyperparms}
\end{table}

\subsection{Training}
\label{appendix:training}
During training we adopted a value of $16$ for batch size and did not use any weight decay. For the BERT + MLP models, we found that they converged within $10$ epochs. For the HDLN model, it took more epoches ($20$ epochs) to converge than the BERT + MLP models did. For all models, we terminated the training when the validation set accuracy no longer improved.
However, since an individual BERT + MLP model was required for each loneliness category, the total training time of the BERT + MLP models were $2.5$ times longer than that of the HDLN model (each individual BERT + MLP model took half the time of HDLN's, but we had to train five BERT + MLP models).

\subsection{Evaluation Metrics}
\label{appendix:results}
% We elaborate the computation of distributional metrics reported in table \ref{tab:eval_fine_grained}.
Denoting a $K$-dimensional target real-value vector as $\boldsymbol{D} = [D_1, D_2, \ldots, D_K]$ and a $K$-dimensional predicted real-value vector as $\hat{\boldsymbol{D}} = [\hat{D}_1, \hat{D}_2, \ldots, \hat{D}_K]$, both satisfy distribution constrains. Accuracy and clark are calculated as in Table~\ref{tab:metrics_calc}.

\begin{table}[ht!]
\centering
% \resizebox{\columnwidth}{!}
{\begin{tabular}{@{}cc@{}}
\toprule
Metric                 & Formulation                                                                             \\ \midrule
Accuracy              & $\mathds{1} \left(arg\,max(\hat{\boldsymbol{D}}) \in arg\,max(\boldsymbol{D})\right)$           \\
Clark                & $\sqrt{\sum_k^K \frac{\left(D_k - \hat{D}_k\right)^2}{\left(D_k + \hat{D}_k\right)^2}}$ \\
% Canberra     & Canberra metric.         & $\sum_k^K \frac{\left\lvert D_k - \hat{D}_k\right\rvert}{D_k + \hat{D}_k}$              \\
% Cosine       & Cosine similarity.       & $\frac{\sum_k^K D_k \hat{D}_k}{\sqrt{\sum_k^K D_k^2}\sqrt{\sum_k^K \hat{D}_k^2}}$       \\
% Intersection & Intersection similarity. & $\sum_k^K \min \left(D_k, \hat{D}_k\right)$                                             
\\ \bottomrule
\end{tabular}}
\caption{Metrics calculation.}
\label{tab:metrics_calc}

\end{table}

\section*{Acknowledgments} \label{sec:acknowledgements}

We thank Terresa Eun and Weier Wan for their helpful discussions and comments. This work was supported by an unrestricted gift from the nonprofit HopeLab Foundation.
LL is generously supported by an Open Philanthropy AI Fellowship.

\bibliographystyle{apalike}
\bibliography{bibliography.bib}

\clearpage
\end{document}